%% file: main.tex
\documentclass[10pt,twocolumn,letterpaper]{article}

\usepackage[pagenumbers]{cvpr} 


\usepackage{xspace}
\usepackage{wrapfig}
\usepackage{tabularx} 
\usepackage{booktabs}
\usepackage{amsmath}
\usepackage{graphicx}
\usepackage{tabularx}
\usepackage{booktabs}
\usepackage{xcolor}
\definecolor{darkgreen}{rgb}{0.0, 0.8, 0.0}
\definecolor{lightred}{rgb}{1, 0.5, 0.5}
\definecolor{lblue}{rgb}{0.38, 0.55, 0.85}
\usepackage{graphicx}

\usepackage{amssymb}
\usepackage{tablefootnote}

\usepackage{enumitem}
\usepackage{colortbl}
\usepackage{arydshln}

\usepackage{multirow}

\usepackage{threeparttable}
\usepackage{float} 

\definecolor{gray}{rgb}{0.85,0.85,0.85} 
\definecolor{lgray}{rgb}{0.96,0.96,0.96} 
\definecolor{dgray}{rgb}{0.5,0.5,0.5} 

\newcommand{\grayline}{\arrayrulecolor{dgray}\hline\arrayrulecolor{black}\arrayrulewidth=3pt}

\definecolor{cvprblue}{rgb}{0.21,0.49,0.74}
\usepackage[pagebackref,breaklinks,colorlinks,allcolors=cvprblue]{hyperref}

\def\paperID{4970} 
\def\confName{CVPR}
\def\confYear{2025}

\title{ProTPS: Prototype-Guided Text Prompt Selection for Continual Learning}

\author{
Jie Mei$^{1}$ \quad 
Li-Leng Peng$^{1}$ \quad 
Keith Fuller$^{2}$ \quad 
Jenq-Neng Hwang$^{1}$ \\
$^{1}$University of Washington, Seattle, WA, USA \\
$^{2}$Alaska Pacific University, Anchorage, Alaska, USA \\
{\tt\small \{jiemei, uwalisha, hwang\}@uw.edu, emsharkproject@gmail.com}
}

\begin{document}
\maketitle

\begin{figure*}[h]
    \centering
    \hfill\raisebox{0.5cm}{CLIP}
    \begin{subfigure}{0.15\linewidth} 
        \centering
        \includegraphics[width=0.5\linewidth]{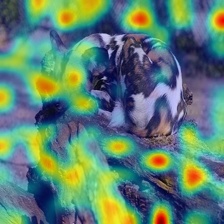}
    \end{subfigure}
    \hspace{-13.5mm}
    \begin{subfigure}{0.15\linewidth} 
        \centering
        \includegraphics[width=0.5\linewidth]{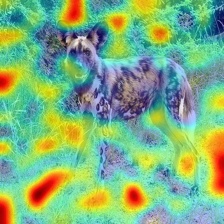}
    \end{subfigure}
    \hspace{-12mm}
    \begin{subfigure}{0.15\linewidth} 
        \centering
        \includegraphics[width=0.5\linewidth]{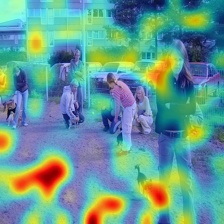}
    \end{subfigure}
    \hspace{-13.5mm}
    \begin{subfigure}{0.15\linewidth} 
        \centering
        \includegraphics[width=0.5\linewidth]{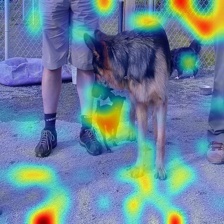}
    \end{subfigure}
    \hspace{-12mm}
    \begin{subfigure}{0.15\linewidth} 
        \centering
        \includegraphics[width=0.5\linewidth]{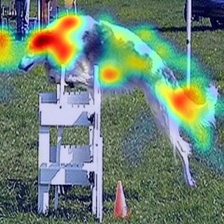}
    \end{subfigure}
    \hspace{-13.5mm}
    \begin{subfigure}{0.15\linewidth} 
        \centering
        \includegraphics[width=0.5\linewidth]{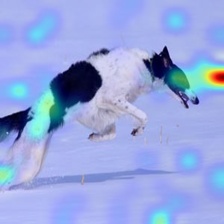}
    \end{subfigure}
    \hspace{-12mm}
    \begin{subfigure}{0.15\linewidth} 
        \centering
        \includegraphics[width=0.5\linewidth]{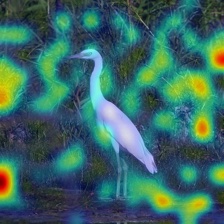}
    \end{subfigure}
    \hspace{-13.5mm}
    \begin{subfigure}{0.15\linewidth} 
        \centering
        \includegraphics[width=0.5\linewidth]{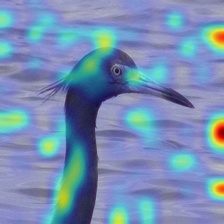}
    \end{subfigure}
    \hspace{-12mm}
    \begin{subfigure}{0.15\linewidth} 
        \centering
        \includegraphics[width=0.5\linewidth]{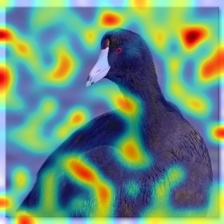}
    \end{subfigure}
    \hspace{-13.5mm}
    \begin{subfigure}{0.15\linewidth} 
        \centering
        \includegraphics[width=0.5\linewidth]{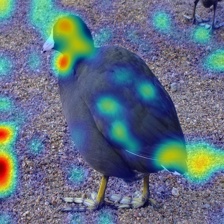}
    \end{subfigure} \\
    
    \hspace{0.03mm}
    \hfill\raisebox{0.5cm}{Img}
    \hspace{0.3mm}
    \begin{subfigure}{0.15\linewidth} 
        \centering
        \includegraphics[width=0.5\linewidth]{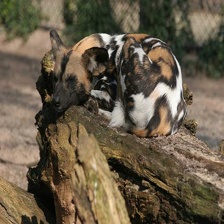}
    \end{subfigure}
    \hspace{-13.5mm}
    \begin{subfigure}{0.15\linewidth} 
        \centering
        \includegraphics[width=0.5\linewidth]{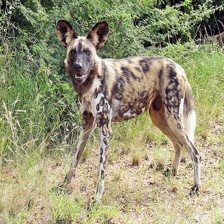}
    \end{subfigure}
    \hspace{-12mm}
    \begin{subfigure}{0.15\linewidth} 
        \centering
        \includegraphics[width=0.5\linewidth]{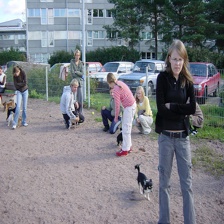}
    \end{subfigure}
    \hspace{-13.5mm}
    \begin{subfigure}{0.15\linewidth} 
        \centering
        \includegraphics[width=0.5\linewidth]{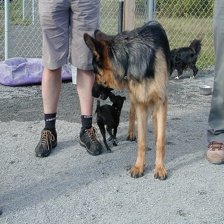}
    \end{subfigure}
    \hspace{-12mm}
    \begin{subfigure}{0.15\linewidth} 
        \centering
        \includegraphics[width=0.5\linewidth]{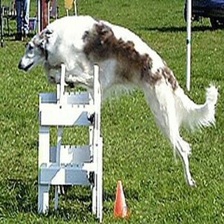}
    \end{subfigure}
    \hspace{-13.5mm}
    \begin{subfigure}{0.15\linewidth} 
        \centering
        \includegraphics[width=0.5\linewidth]{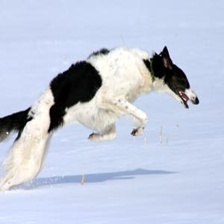}
    \end{subfigure}
    \hspace{-12mm}
    \begin{subfigure}{0.15\linewidth} 
        \centering
        \includegraphics[width=0.5\linewidth]{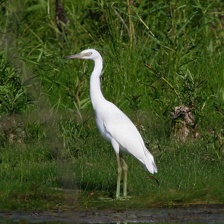}
    \end{subfigure}
    \hspace{-13.5mm}
    \begin{subfigure}{0.15\linewidth} 
        \centering
        \includegraphics[width=0.5\linewidth]{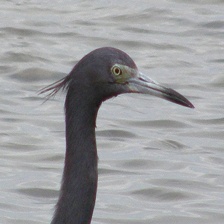}
    \end{subfigure}
    \hspace{-12mm}
    \begin{subfigure}{0.15\linewidth} 
        \centering
        \includegraphics[width=0.5\linewidth]{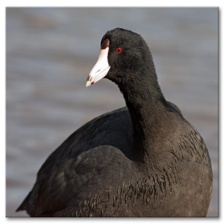}
    \end{subfigure}
    \hspace{-13.5mm}
    \begin{subfigure}{0.15\linewidth} 
        \centering
        \includegraphics[width=0.5\linewidth]{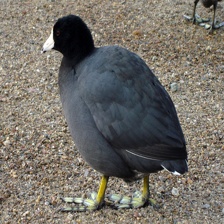}
    \end{subfigure}\\

    \hfill\raisebox{0.5cm}{Ours}
    \hspace{-0.1mm}
    \begin{subfigure}{0.15\linewidth} 
        \centering
        \includegraphics[width=0.5\linewidth]{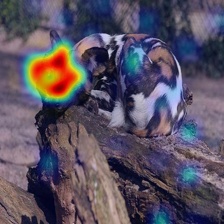}
    \end{subfigure}
    \hspace{-13.5mm}
    \begin{subfigure}{0.15\linewidth} 
        \centering
        \includegraphics[width=0.5\linewidth]{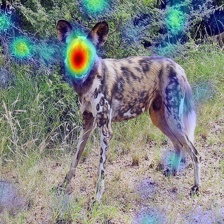}
    \end{subfigure}
    \hspace{-12mm}
    \begin{subfigure}{0.15\linewidth} 
        \centering
        \includegraphics[width=0.5\linewidth]{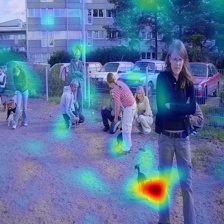}
    \end{subfigure}
    \hspace{-13.5mm}
    \begin{subfigure}{0.15\linewidth} 
        \centering
        \includegraphics[width=0.5\linewidth]{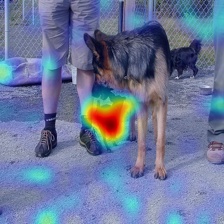}
    \end{subfigure}
    \hspace{-12mm}
    \begin{subfigure}{0.15\linewidth} 
        \centering
        \includegraphics[width=0.5\linewidth]{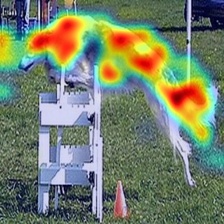}
    \end{subfigure}
    \hspace{-13.5mm}
    \begin{subfigure}{0.15\linewidth} 
        \centering
        \includegraphics[width=0.5\linewidth]{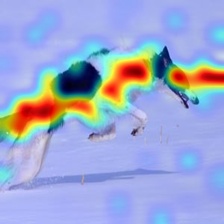}
    \end{subfigure}
    \hspace{-12mm}
    \begin{subfigure}{0.15\linewidth} 
        \centering
        \includegraphics[width=0.5\linewidth]{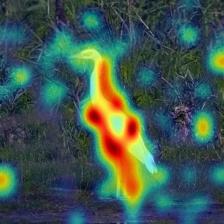}
    \end{subfigure}
    \hspace{-13.5mm}
    \begin{subfigure}{0.15\linewidth} 
        \centering
        \includegraphics[width=0.5\linewidth]{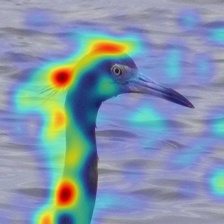}
    \end{subfigure}
    \hspace{-12mm}
    \begin{subfigure}{0.15\linewidth} 
        \centering
        \includegraphics[width=0.5\linewidth]{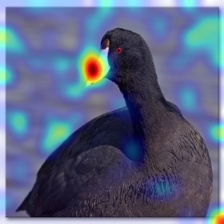}
    \end{subfigure}
    \hspace{-13.5mm}
    \begin{subfigure}{0.15\linewidth} 
        \centering
        \includegraphics[width=0.5\linewidth]{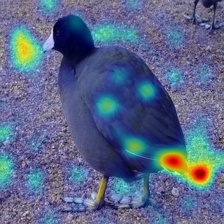}
    \end{subfigure}\\

    \caption{Visualization of learned text prompts via Grad-CAM~\cite{selvaraju2017grad}. Highlighted image areas are the attended regions of the learned text prompt of each class. Each pair of adjacent columns, moving from left to right, represents the same class: ``african hunting dog'', ``chihuahua'', ``borzoi'', ``little blue heron'', ``american coot''.}
    \label{fig: prompt vis}
\end{figure*}

\input{sec/0_abstract}

\input{sec/1_intro}    
\input{sec/2_relate}

\input{sec/3_method}

\input{sec/4_exp}

\input{sec/5_conclu}

\newpage

{
    \small
    \bibliographystyle{ieeenat_fullname}
    \bibliography{cvpr2025}
}

\input{sec/X_suppl}

\end{document}

%% file: sec/0_abstract.tex
\begin{abstract}
For continual learning, text-prompt-based methods leverage text encoders and learnable prompts to encode semantic features for sequentially arrived classes over time. A common challenge encountered by existing works is how to learn unique text prompts, which implicitly carry semantic information of new classes, so that the semantic features of newly arrived classes do not overlap with those of trained classes, thereby mitigating the catastrophic forgetting problem. To address this challenge, we propose a novel approach ``\textbf{Pro}totype-guided \textbf{T}ext \textbf{P}rompt \textbf{S}election (ProTPS)'' to intentionally increase the training flexibility thus encouraging the learning of unique text prompts. Specifically, our ProTPS learns class-specific vision prototypes and text prompts. Vision prototypes guide the selection and learning of text prompts for each class. We first evaluate our ProTPS in both class incremental (CI) setting and cross-datasets continual (CDC) learning setting. Because our ProTPS achieves performance close to the upper bounds, we further collect a real-world dataset with 112 marine species collected over a span of six years, named Marine112, to bring new challenges to the community. Marine112 is authentically suited for the class and domain incremental (CDI) learning setting and is under natural long-tail distribution. The results under three settings show that our ProTPS performs favorably against the recent state-of-the-art methods. The implementation code and Marine112 dataset will be released upon the acceptance of our paper.
\end{abstract}

%% file: sec/1_intro.tex
\section{Introduction}
\label{sec:intro}

\begin{figure*}[htbp]
    \centering
    \includegraphics[width=2\columnwidth]{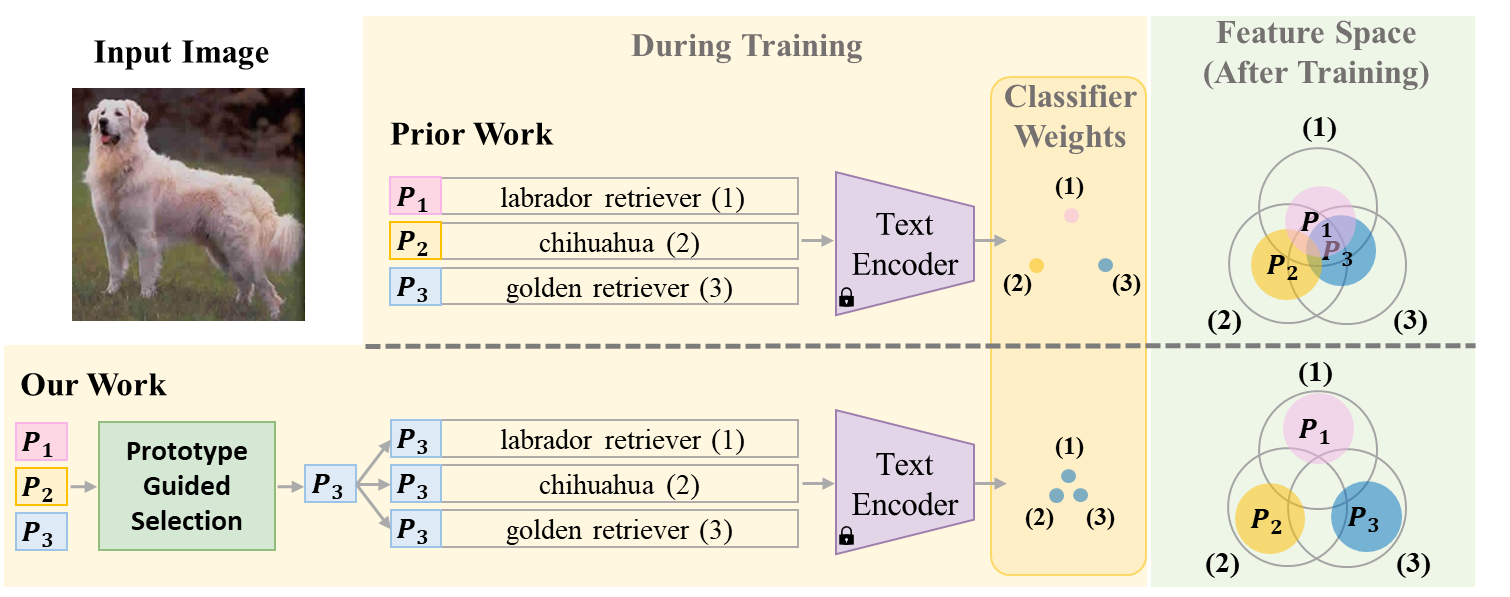}
    \caption{\textbf{Prior work}~\cite{zhou2022learning} proposes learnable prompts to replace predefined prompt templates. \textbf{Our work} introduces a novel text prompt selection approach that enables semantic prompts of each class to interact with all other class names. Note that three big gray circles in the feature space represent all possible semantic features for each class. 
    }
    \label{fig:intro_comparison}
\end{figure*}

The dynamic nature of real-world data, with its constant evolution and shifting paradigms, such as the incremental addition of new classes, poses significant challenges. When a model's access to historical training data is constrained, due to privacy considerations, data storage and transfer limitations, or energy efficiency concerns, it can only be primarily updated with new data. This scenario leads to a deep learning model's ``catastrophic forgetting'' of previously learned knowledge, a critical problem documented in seminal works~\cite{rebuffi2017icarl, kirkpatrick2017overcoming, wu2019large, li2017learning, wang2022learning, wang2023attriclip, jung2023generating, mei2022hcil} in the area of continual learning.

To mitigate the catastrophic forgetting problem, recent text-prompt-based methods~\cite{wang2023attriclip, zhou2022learning, wang2022s, thengane2022continualclip} leverage frozen Contrastive Language-Image Pre-training (CLIP)~\cite{radford2021learning}. In these methods, text prompts are concatenated with class names as input to the frozen pre-trained CLIP~\cite{radford2021learning} text encoder to get semantic features of each class, which then serve as weights of the linear classifier as proposed in the original CLIP~\cite{radford2021learning}. These methods use learnable or predefined prompt to encode semantic features for sequentially arrived classes over time. However, \textbf{a common challenge} encountered by existing works is how to learn text prompts encoded with unique semantic features of different classes so that the semantic features of newly arrived classes do not overlap with those of trained classes.

To address this challenge, we propose a novel approach Prototype-guided Text Prompt Selection (ProTPS). Our \textbf{motivation} is to intentionally increase the training flexibility thus encouraging the text prompts to encode unique semantic features of each class. For the toy example in Fig.~\ref{fig:intro_comparison}, given an input image of the golden retriever in training, the text prompt belonging to ``golden retriever'', i.e., $P_3$, can only interact with ``golden retriever'' in the prior work, while our ProTPS selects $P_3$ to concatenate with all class names, increasing the training flexibility. By doing so, our prompts ``$P_3$-labrador retriever'' and ``$P_3$-golden retriever'' are closer in the feature space at the beginning of training, compared with ``$P_1$-labrador retriever'' and ``$P_3$-golden retriever'' in the prior work. Then using semantic features of our prompts as the linear classifier weights results in a bigger training loss. By minimizing the loss, our $P_3$ is encouraged to capture the unique semantic features of ``golden retriever''. After training, our class-specific text prompts can be more distinct in the feature space as illustrated in Fig.~\ref{fig:intro_comparison}.




\textbf{Why use vision prototypes as guidance?} Different from existing text-prompts-based works which have a single classifier derived from the text encoder as proposed in the CLIP~\cite{radford2021learning}, our ProTPS has an additional linear classifier whose weights are initialized with visual features, i.e., prototypes, of each class, computed from the pre-trained image encoder. At the beginning of training, this linear classifier can provide a reasonable initial prediction. Our \textbf{motivation} for using prototypes as guidance for text prompt selection stems from the hypothesis that vision prototypes and text prompts can be complementary features, i.e., vision prototypes can be good at globally distinct categories, such as cows and birds; on the other hand, text prompts can be trained to capture unique semantic features corresponding to regional image details such as face, body, tail, etc. 




The experimental configurations for existing continual learning methods are idealistic, typically partitioning a single dataset into multiple non-overlapped and data-balanced tasks, i.e., class incremental (CI) learning, or combining two datasets into a long-sequence domain-shift task, i.e., cross-datasets continual (CDC) learning. To handle a more realistic scenario, we collect a real-world challenging dataset, named Marine112, covering 112 marine species across 6 years. We believe Marine112 is of great importance to the continual learning community, for the following reasons: (1) it features overlapped unique classes among different tasks with varied imaging qualities, i.e., domain shift; (2) it reflects the longtail~\cite{liu2019large} (imbalanced) distribution observed in nature; and (3) it is naturally suited for the class and domain incremental (CDI) learning. Our contributions are summarized as follows:

\begin{itemize}
  \item We propose a novel approach ProTPS to continually learn \textbf{unique} text prompts so that the semantic features of newly arrived classes do not overlap with those of trained classes, thereby mitigating ``catastrophic forgetting''.
  \item Our novel ProTPS learns global image-level vision prototypes to guide the selection and learning of text prompts that attend to \textbf{region-level} image details.
  \item In class incremental learning setup, ProTPS attains +$1.9\%$ and +$10.6\%$ gains on CIFAR100 and ImageNet100 respectively over the current state-of-the-art methods. In cross-datasets continual learning setup, ProTPS secures +$2.1\%$ and +$7.3\%$ on ``ImageNet2CIFAR'' and ``ImageNet+CIFAR'' respectively.
  \item We collect a real-world Marine112 dataset, with long-tail distribution and authentically well-suited for the class and domain incremental (CDI) learning task. It brings new challenges to the continual learning community.
\end{itemize}

%% file: sec/2_relate.tex
\section{Related Work}
\label{sec: related}

\textbf{Classic methods.} Rehearsal-based continual learning~\cite{douillard2021dytox, rebuffi2017icarl, hou2019learning, castro2018end, robins1995catastrophic} is highly effective by keeping a fixed amount of previously trained data as a memory. Yet, with potential privacy issues~\cite{shokri2015privacy}, they also tend to underperform as the size or quality of the rehearsal memory decreases. Dynamic architectures~\cite{yoon2017lifelong, li2019learn, hung2019compacting, fernando2017pathnet, golkar2019continual, serra2018overcoming, wu2022class} expand classification heads and/or backbones for new training tasks. Unfortunately, at inference time, some dynamic architectures need complex pruning~\cite{golkar2019continual} as post-processing or require an impractical task identifier~\cite{fernando2017pathnet, serra2018overcoming, hung2019compacting}.

\textbf{Text-prompt-based methods.} Continual-CLIP~\cite{thengane2022continualclip} directly applies the pre-trained frozen CLIP~\cite{radford2021learning} to the continual learning scenario by concatenating class names with a single text prompt template, \eg, ``a bad photo of a \{ \}'', to derive the classifier. Thus, without new parameters, its performance is capped by the pre-trained CLIP~\cite{radford2021learning}. To replace predefined templates, CoOP~\cite{zhou2022learning}, built upon CLIP~\cite{radford2021learning}, proposes learnable class-specific prompts but its prompts cannot interact among different classes. AttriCLIP~\cite{wang2023attriclip} boosts the performance with an attribute word bank which is shared for all classes, and is continually trained on newly arrived classes. Inevitably its word bank may overfit attributes of newly arrived classes. In contrast, our novel ProTPS continually encourages text prompts to capture class-specific \textbf{unique} features so that the semantic features of newly arrived classes do not overlap with those of trained classes.

\textbf{Visual-prompt-based methods.} Recent visual prompt tuning works~\cite{wang2022dualprompt, li2023learning, wang2022learning, smith2023coda, wang2023hide, tang2023prompt}, built upon a pre-trained image encoder, employ a dynamic query mechanism based on key-value pairs to learn instance-specific prompts tailored to the context. The L2P framework~\cite{wang2022learning} pioneers the amalgamation of visual prompts with continual learning by advocating for a universal prompt repository to facilitate task adaptation. DualPrompt~\cite{wang2022dualprompt} further innovates by affixing task-invariant and task-specific visual prompts to the pre-trained image encoder. CODA-P~\cite{smith2023coda} introduces decomposed and expanded visual prompts that can be optimized in an end-to-end fashion. HiDe-Prompt~\cite{wang2023hide} explicitly optimizes three hierarchical components for visual prompt tuning. However, unlike text-prompt-based methods, these methods only benefit from pre-trained image encoders.

\begin{figure*}[t]
    \centering
    \includegraphics[width=2\columnwidth]{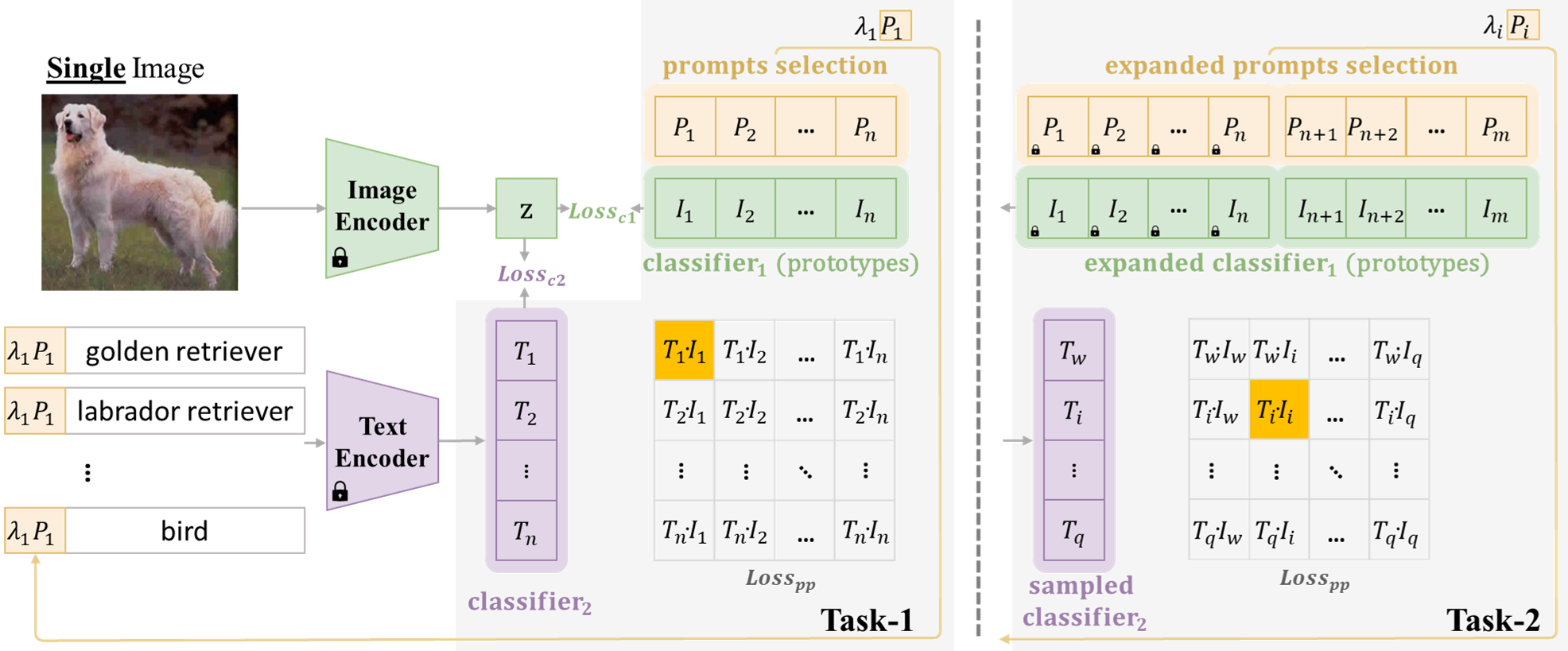}
    \caption{\textbf{ProTPS framework.} In Task-1, we use the prototype $I_i$ of class $i$ to initialize the weights of the linear $classifier_{1}$, which is referred to as ProTPS's ``vision classifier''. We finetune these vision prototypes for refinement. Each class prototype $I_i$ is paired with a learnable text prompt $P_i$. For a single input image, we select a text prompt to concatenate with all class names in Task-1 to create $classifier_{2}$, which is referred to as the ``text classifier'' of ProTPS. In Task-2, prototypes and paired prompts are expanded. We freeze prototypes $I_1, ..., I_n$ and prompts $P_1, ..., P_n$ trained in the previous task. We propose a sampling method to create $sampled$ $classifier_{2}$ to help new text prompts capture exclusive unique features for new classes. For the later tasks, the expansion rule is the same as Task-2.}
    \label{fig:piepline}
\end{figure*}



\textbf{Prototype learning.} Prototype learning aims to learn prototype vectors to represent each class. Classic continual learning methods like iCaRL~\cite{rebuffi2017icarl}, LUCIR~\cite{hou2019learning}, PODnet~\cite{douillard2020podnet} and DER~\cite{yan2021dynamically} rely on the nearest-prototype classifier~\cite{webb2011statistical, mensink2012metric}. Recent methods~\cite{zhu2021prototype, zhu2022self,toldo2022bring} use prototypes to generate synthetic samples of trained classes or use synthetic samples to calculate prototypes~\cite{wang2022beef}. In contrast, we hypothesize that prototypes derived from the image encoder are mainly global image-level representations. Thus, we propose to jointly learn unique text prompts capturing exclusive semantic features of each class.

\textbf{Contrastive learning.}
Recently, the contrastive learning field has shown notable progress within self-supervised representation learning~\cite{wu2018unsupervised, henaff2020data, oord2018representation, tian2020contrastive, hjelm2018learning}. SupCon~\cite{khosla2020supervised} extends the self-supervised batch contrastive approach to the supervised setting, allowing the model to leverage label information to construct many positives per anchor in addition to many negatives. CLIP~\cite{radford2021learning} applies supervised contrastive learning between images and text pairs to align them to a common feature space. These methods are designed to enhance the feature extraction ability of encoders.


%% file: sec/3_method.tex
\section{Methodology}




\label{sec: pilot_study}

\textbf{Continual learning formulation.} We define a sequence of $T$ training tasks, denoted by $D=
\{D_1, . . . , D_T\}$, with each task $D_t =\{(x_{i}^{t}, y_{i}^{t})\}_{i=1}^{n_t}$ comprising $n_t$ images $x_{i}^{t}$ and their respective labels $y_{i}^{t}$. Training on task $D_t$ occurs without access to data from prior tasks $\{D_1, . . . , D_{t-1}\}$. In the conventional class incremental learning scenario, each task contains non-overlapped classes.

\subsection{Framework of ProTPS}
\label{sec: Framework}
 Our proposed novel Prototype-guided Text Prompt Selection (ProTPS) is built upon frozen CLIP~\cite{radford2021learning}. In contrast to existing text-prompt-based methods~\cite{zhou2022learning, wang2023attriclip, thengane2022continualclip, wang2022s}, our ProTPS continually encourages text prompts to capture class-specific unique semantic features.

\textbf{Refinement of vision prototypes.} We use the prototype $I_i$ of class $i$ to initialize the weights of the $classifier_{1}$ at the beginning of Task-1  as shown in Fig.~\ref{fig:piepline}, which is referred to as ProTPS's ``vision classifier''. We can see vision prototypes as global image-level representations of each class. Throughout the training of Task-1, we finetune these vision prototypes for refinement. As new tasks commence, while the ``vision classifier'' is expanded with prototypes of new classes, we ensure previously learned prototypes remain frozen to conserve the integrity of earlier learned representations, as illustrated in Fig.~\ref{fig:piepline}. 

\textbf{Prototypes-guided text prompts selection.} Each class prototype $I_i$ is paired with a learnable text prompt $P_i$. For a single image, ProTPS selects the text prompt that corresponds to the highest cosine similarity score $\lambda_i$ as indicated by vision prototypes. Recognizing that the ``vision classifier'' may not always yield accurate prediction, we scale the selected text prompt by $\lambda_i$ as shown in Fig.~\ref{fig:piepline} and concatenate it at the embedding level with all class names in Task-1. These combined textual inputs are processed through the text encoder to derive $classifier_{2}$, which is referred to as ProTPS's ``text classifier'' and can be seen as \textbf{unique} semantic features of each class.

When new tasks start, we expand text prompts for new classes and ensure previously learned prompts remain frozen to conserve the integrity of earlier learned unique semantic features, as depicted in Fig.~\ref{fig:piepline}. Besides new class names, we randomly sample a fixed number of class names from previous classes before concatenating with the selected text prompt in the new task, yielding $sampled$ $classifier_2$ in Fig.~\ref{fig:piepline}. This sampling strategy allows the newly expanded text prompts to interact with previous class names to help new text prompts learn unique semantic features for new classes while keeping a constant GPU memory usage.

\textbf{One aggregated classifier.} During inference, we also keep the proposed prompt selection and ProTPS generates predictions by calculating the average cosine similarity score of each class from its ``vision classifier'' and ``text classifier''.

\subsection{Optimization Objectives of ProTPS}
\label{sec: loss_functions}

\textbf{Classification losses.} Given one single image input of class $i$, we first compute two cross-entropy losses respectively for dual classifiers of ProTPS based on the image feature $\mathbf{z}$:
\begin{equation}
\label{eq: l_c1}
Loss_{c1} = -\log \frac{e^{\left<\mathbf{z},\ \mathbf{I}_i\right>/\tau}}{\sum_{k=k_0}^{k_1} e^{\left<\mathbf{z},\ \mathbf{I}_k\right>/\tau}},
\end{equation}
\begin{equation}
Loss_{c2} = -\log \frac{e^{\left<\mathbf{z},\ \mathbf{T}_i\right>/\tau}}{\sum_{k=k'_0}^{k'_1} e^{\left<\mathbf{z},\ \mathbf{T}_k\right>/\tau}},
\end{equation}
where $\tau$ is the temperature parameter learned by CLIP~\cite{radford2021learning}, $\left<\cdot,\cdot\right>$ denotes the cosine similarity, $k_0$ and $k_1$ respectively represent the start and end indices of newly expanded prototypes by default unless otherwise specified in experiments, \eg, $k_0=n+1$ and $k_1=m$ in Task-2 as depicted in Fig.~\ref{fig:piepline}. This is to ensure that one of the newly expanded text prompts can be selected and optimized to learn unique features of new classes. $k'_0$ and $k'_1$ respectively represent the start and end indices of weights vectors in the ``text classifier'', \eg, $k'_0=1$ and $k'_1=n$ in Task-1 and $k'_0=w$ and $k'_1=q$ in Task-2 as depicted in Fig.~\ref{fig:piepline}.

\textbf{Prompt-prototype contrastive loss.} Although vision prototypes and text prompts represent respectively global image-level representations and unique semantic features of each class, we propose a ``prompt-prototype contrastive loss'', $Loss_{pp}$, to align and enhance text prompts and vision prototypes, which further boosts the performance. It is different from the well-studied contrastive loss, such as SupCon~\cite{khosla2020supervised}, in two aspects: (1) it refocuses the goal from improving encoders to refining text prompts and vision prototypes, and (2) it is formulated using just a single input image. Given one single image input of class $i$, the proposed ``prompt-prototype contrastive loss'', depicted in Fig.~\ref{fig:piepline}, is as follows:
\begin{equation}
Loss_{pp} = (1 - \langle \mathbf{T}_i,\ \mathbf{I}_i \rangle) + \frac{1}{N} \sum_{\substack{j \neq i \text{ or } k \neq i}}^N \langle \mathbf{T}_j,\ \mathbf{I}_k \rangle,
\label{eq: loss_cc}
\end{equation}
where $\left<\cdot,\cdot\right>$ denotes the cosine similarity. The first term is the cosine distance of the positive pair (yellow colored grids in Fig.~\ref{fig:piepline}) between prototypes and weights of ``text classifier'', given the single input image of class $i$; the second term is the mean cosine similarity of negative pairs (uncolored grids in Fig.~\ref{fig:piepline}) between prototypes and weights of ``text classifier''. $N$ is the total number of negative pairs. With $\lambda_{pp}$ as the balance factor, the whole optimization objective is:
\begin{equation}
Loss = Loss_{c1} + Loss_{c2} +\lambda_{pp}\cdot Loss_{pp},
\end{equation}

\begin{table*}
    \setlength{\tabcolsep}{7pt}
    \centering
    \begin{tabularx}{\textwidth}{cccccccccccc}
        \toprule
        Prompt & Method  & T-1 & T-2 & T-3 & T-4 & T-5 & T-6 & T-7 & T-8 & T-9 & T-10 \\
        \midrule
       \multirow{4}{*}{$visual$} & \cellcolor{lgray}L2P~\cite{wang2022learning}  & \cellcolor{lgray}60.9 & \cellcolor{lgray}53.5 & \cellcolor{lgray}49.8 & \cellcolor{lgray}48.6 & \cellcolor{lgray}47.2 & \cellcolor{lgray}48.1 & \cellcolor{lgray}49.5 &\cellcolor{lgray}49.7 & \cellcolor{lgray}48.1 & \cellcolor{lgray}48.3 \\
        & \cellcolor{lgray}DualPrompt~\cite{wang2022dualprompt} & \cellcolor{lgray}72.5 & \cellcolor{lgray}60.8 & \cellcolor{lgray}55.7 & \cellcolor{lgray}54.3 & \cellcolor{lgray}53.2 & \cellcolor{lgray}53.6 & \cellcolor{lgray}53.8 &\cellcolor{lgray}53.3 & \cellcolor{lgray}52.8 & \cellcolor{lgray}52.6 \\
        & \cellcolor{lgray}CODA-P~\cite{smith2023coda} & \cellcolor{lgray}91.1 & \cellcolor{lgray}86.5 & \cellcolor{lgray}82.9 & \cellcolor{lgray}80.3 & \cellcolor{lgray}79.8 & \cellcolor{lgray}78.2 & \cellcolor{lgray}77.1 & \cellcolor{lgray}77.2 &\cellcolor{lgray}75.6 & \cellcolor{lgray}75.2 \\
        & \cellcolor{lgray}HiDe-Prompt~\cite{wang2023hide}  & \cellcolor{lgray}92.3 & \cellcolor{lgray}\underline{90.1} & \cellcolor{lgray}\underline{89.8} & \cellcolor{lgray}\underline{87.5} & \cellcolor{lgray}\underline{85.5} & \cellcolor{lgray}85.9 & \cellcolor{lgray}84.5 & \cellcolor{lgray}\underline{85.9} & \cellcolor{lgray}83.5 & \cellcolor{lgray}82.9 \\
        \cline{1-1}
        \arrayrulecolor{gray}\cline{1-1}\arrayrulecolor{black}
        \multirow{4}{*}{$text$} & \cellcolor{gray}CoOp~\cite{zhou2022learning} (1000) & \cellcolor{gray}89.2 &\cellcolor{gray}\cellcolor{gray}83.2 & \cellcolor{gray}76.7 & \cellcolor{gray}79.8 &\cellcolor{gray}79.9 & \cellcolor{gray}82.3 & \cellcolor{gray}79.7 & \cellcolor{gray}80.1 & \cellcolor{gray}80.3 & \cellcolor{gray}79.3 \\
        & \cellcolor{gray}Continual-CLIP~\cite{thengane2022continualclip}  & \cellcolor{gray}93.3 & \cellcolor{gray}87.6 & \cellcolor{gray}83.1 & \cellcolor{gray}81.7 & \cellcolor{gray}80.5 & \cellcolor{gray}80.2 & \cellcolor{gray}79.3 & \cellcolor{gray}78.5 & \cellcolor{gray}76.9 & \cellcolor{gray}75.4 \\
        & \cellcolor{gray}Zero-Shot CLIP~\cite{radford2021learning}  & \cellcolor{gray}91.8 & \cellcolor{gray}89.2 & \cellcolor{gray}89.3 & \cellcolor{gray}91.1 & \cellcolor{gray}92.2 & \cellcolor{gray}92.6 & \cellcolor{gray}90.9 & \cellcolor{gray}79.3 & \cellcolor{gray}78.5 & \cellcolor{gray}75.1 \\
        & A\cellcolor{gray}ttriCLIP~\cite{wang2023attriclip}& \cellcolor{gray}\underline{95.4} & \cellcolor{gray}89.4 & \cellcolor{gray}84.5 &\cellcolor{gray}86.7 & \cellcolor{gray}84.4 & \cellcolor{gray}\underline{86.6} & \cellcolor{gray}\underline{85.9} & \cellcolor{gray}85.6 & \cellcolor{gray}\underline{86.9} & \cellcolor{gray}\underline{83.3} \\
        \midrule
        
        
        $text^*$ &  \textbf{ProTPS} (ours)  & \textbf{95.5} & \textbf{93.4} & \textbf{93.7} & \textbf{94.0} & \textbf{94.7} & \textbf{95.3} & \textbf{94.7} & \textbf{94.2} & \textbf{93.8} & \textbf{93.9}\textcolor{darkgreen}{(+10.6)} \\ 
        \midrule
        & Upper bound  & - & - & - & - & - & - & - & - & - & 95.5 \\
        \bottomrule
    \end{tabularx}
    \caption{Accuracy of different continual learning methods on ImageNet100~\cite{deng2009imagenet} under CI setting. Second-best results are underlined. T-i denotes Task-i.}
    \label{exp: imagenet100}
\end{table*}

%% file: sec/4_exp.tex
\section{Experiments}
\label{sec:exp}

In this section, we first detail the implementation. We then compare the proposed ProTPS with other popular methods in the class-incremental (CI) setting and the cross-datasets continual (CDC) learning setting. We also introduce a new real-world dataset, Marine112, for class and domain incremental (CDI) learning setting. Finally, we present ablation studies to evaluate ProTPS's individual components.

\begin{table*}
    \setlength{\tabcolsep}{7pt}
    \centering
    \begin{tabularx}{\textwidth}{cccccccccccc}
        \toprule
        Prompt & Method  & T-1 & T-2 & T-3 & T-4 & T-5 & T-6 & T-7 & T-8 & T-9 & T-10 \\
        \midrule
        \multirow{4}{*}{$visual$} & \cellcolor{lgray}L2P~\cite{wang2022learning}  &\cellcolor{lgray}60.1 & \cellcolor{lgray}47.8 & \cellcolor{lgray}45.9 & \cellcolor{lgray}47.1 & \cellcolor{lgray}48.9 & \cellcolor{lgray}48.9 & \cellcolor{lgray}49.4 & \cellcolor{lgray}49.0 & \cellcolor{lgray}50.0 & \cellcolor{lgray}49.4 \\
          & \cellcolor{lgray}DualPrompt~\cite{wang2022dualprompt}  & \cellcolor{lgray}73.9 & \cellcolor{lgray}60.2 &\cellcolor{lgray}56.4 & \cellcolor{lgray}55.3 & \cellcolor{lgray}54.5 & \cellcolor{lgray}53.3 & \cellcolor{lgray}53.4 &\cellcolor{lgray}53.1 & \cellcolor{lgray}53.5 &\cellcolor{lgray}54.0 \\
           & \cellcolor{lgray}CODA-P~\cite{smith2023coda}  & \cellcolor{lgray}94.3 & \cellcolor{lgray}87.3 &\cellcolor{lgray}83.7 & \cellcolor{lgray}82.7 & \cellcolor{lgray}81.3 & \cellcolor{lgray}78.2 & \cellcolor{lgray}76.7 & \cellcolor{lgray}74.0 & \cellcolor{lgray}74.1 &\cellcolor{lgray}72.3 \\
          & \cellcolor{lgray}HiDe-Prompt~\cite{wang2023hide}  & \cellcolor{lgray}95.4 & \cellcolor{lgray}93.6 & \cellcolor{lgray}90.7 & \cellcolor{lgray}\underline{89.5} & \cellcolor{lgray}\cellcolor{lgray}\underline{87.9} &  \cellcolor{lgray}\underline{85.5} & \cellcolor{lgray} \underline{84.0} & \cellcolor{lgray}\underline{82.0} &  \cellcolor{lgray}\underline{82.2} & \cellcolor{lgray}81.0 \\
        \cline{1-1}
        \arrayrulecolor{gray}\cline{1-1}\arrayrulecolor{black}
        \multirow{4}{*}{$text$} & \cellcolor{gray}CoOp~\cite{zhou2022learning} (1000) & \cellcolor{gray}95.8 & \cellcolor{gray}90.7 & \cellcolor{gray}85.2 & \cellcolor{gray}83.4 & \cellcolor{gray}80.8 & \cellcolor{gray}75.8 & \cellcolor{gray}74.7 & \cellcolor{gray}71.7 &\cellcolor{gray}\cellcolor{gray}71.3 &\cellcolor{gray}67.6 \\
        & \cellcolor{gray}Continual-CLIP~\cite{thengane2022continualclip}  & \cellcolor{gray}95.3 & \cellcolor{gray}92.3 &\cellcolor{gray}82.1 & \cellcolor{gray}79.2 & \cellcolor{gray}78.5 &\cellcolor{gray}77.9 & \cellcolor{gray}75.7 &\cellcolor{gray}75.2 & \cellcolor{gray}74.8 & \cellcolor{gray}73.4 \\
        & \cellcolor{gray}Zero-Shot CLIP~\cite{radford2021learning} & \cellcolor{gray}95.7 & \cellcolor{gray}92.9 & \cellcolor{gray}85.2 & \cellcolor{gray}82.9 & \cellcolor{gray}82.3 & \cellcolor{gray}82.1 & \cellcolor{gray}79.6 & \cellcolor{gray}79.6 & \cellcolor{gray}79.2 & \cellcolor{gray}78.3 \\
        & \cellcolor{gray}AttriCLIP~\cite{wang2023attriclip}  & \cellcolor{gray}\textbf{97.8} &\cellcolor{gray}\underline{93.7} &\cellcolor{gray}\underline{91.0} & \cellcolor{gray}87.5 &\cellcolor{gray}84.7 & \cellcolor{gray}82.5 &\cellcolor{gray}82.3 & \cellcolor{gray}81.9 & \cellcolor{gray}81.7 & \cellcolor{gray}\underline{81.4} \\
        \midrule

         
         $text^*$& \textbf{ProTPS} (ours)  & \textbf{97.8} & \textbf{94.5} & \textbf{91.6} & \textbf{90.6} & \textbf{89.1} & \textbf{87.3} & \textbf{85.5} & \textbf{84.3} & \textbf{84.2} & \textbf{83.3}\textcolor{darkgreen}{(+1.9)} \\ 
        \midrule
         & Upper bound & - & - & - & - & - & - & - & - & - & 86.7 \\
        \bottomrule
    \end{tabularx}
    \caption{Accuracy of different continual learning methods on CIFAR100~\cite{krizhevsky2009learning} under CI setting. Second-best results are underlined. T-i denotes Task-i.}
    \label{exp: cifar100}
\end{table*}

\textbf{Datasets.} Our experiments are conducted on ImageNet100, CIFAR100~\cite{lopez2017gradient}, and Marine112. ImageNet100 is a subset of ImageNet-1K~\cite{deng2009imagenet}, containing images sized 224x224 from 100 classes. We use the same 100 classes as AttriCLIP~\cite{wang2023attriclip}, with about 1,300 training and 50 testing images per class. Details are provided in the Appendix~\ref{sec: Classes of ImageNet100}. CIFAR100 has 100 classes with 500 training and 100 testing images per class. Both CIFAR100 and ImageNet100 are split into 10 tasks with 10 classes in each task. Besides, we collect a real-world Marine112 dataset, which is naturally suited for the class and domain incremental (CDI) learning and reflects the long-tail distribution~\cite{liu2019large} observed in nature. Its data covers 112 marine species in total across 6 different years. 

\textbf{Metric.} After learning each task $T-i$, the accuracy is evaluated on exposed classes of all trained tasks (i.e., $T-1, 2, ..., i$)~\cite{wang2023attriclip}, as detailed in Appendix~\ref{sec: metric}.


\textbf{Baselines.} We compare the proposed ProTPS with (1) \textbf{visual-prompt-based methods} (L2P~\cite{wang2022learning}, DualPrompt~\cite{wang2022dualprompt}, CODA-P~\cite{smith2023coda}, and HiDe-Prompt~\cite{wang2023hide}), and (2) \textbf{text-prompt-based methods} (CoOp~\cite{zhou2022learning}, the only compared method with 1000 memory buffer, ContinualCLIP~\cite{thengane2022continualclip}, Zero-Shot CLIP~\cite{radford2021learning}, and AttriCLIP~\cite{wang2023attriclip}). Compared with AttriCLIP~\cite{wang2023attriclip}, our ProTPS introduces a negligible memory increase of 105 KB, i.e., 0.006\% of 1.71 GB full model size. The upper-bound method is the Linear Probe CLIP from~\cite{radford2021learning} by using all tasks' training data at the same time to optimize a linear classifier. We use the original CLIP pre-trained ViT-L/14~\cite{dosovitskiy2020image} as the image encoder for the upper-bound method, our ProTPS and all baselines (explanation and training details are in Appendix~\ref{sec: ProTPS training details}).

\subsection{Class-Incremental Learning}
\label{sec: CIL}

\begin{figure}[htbp]
    \centering
    \includegraphics[width=0.5\textwidth]{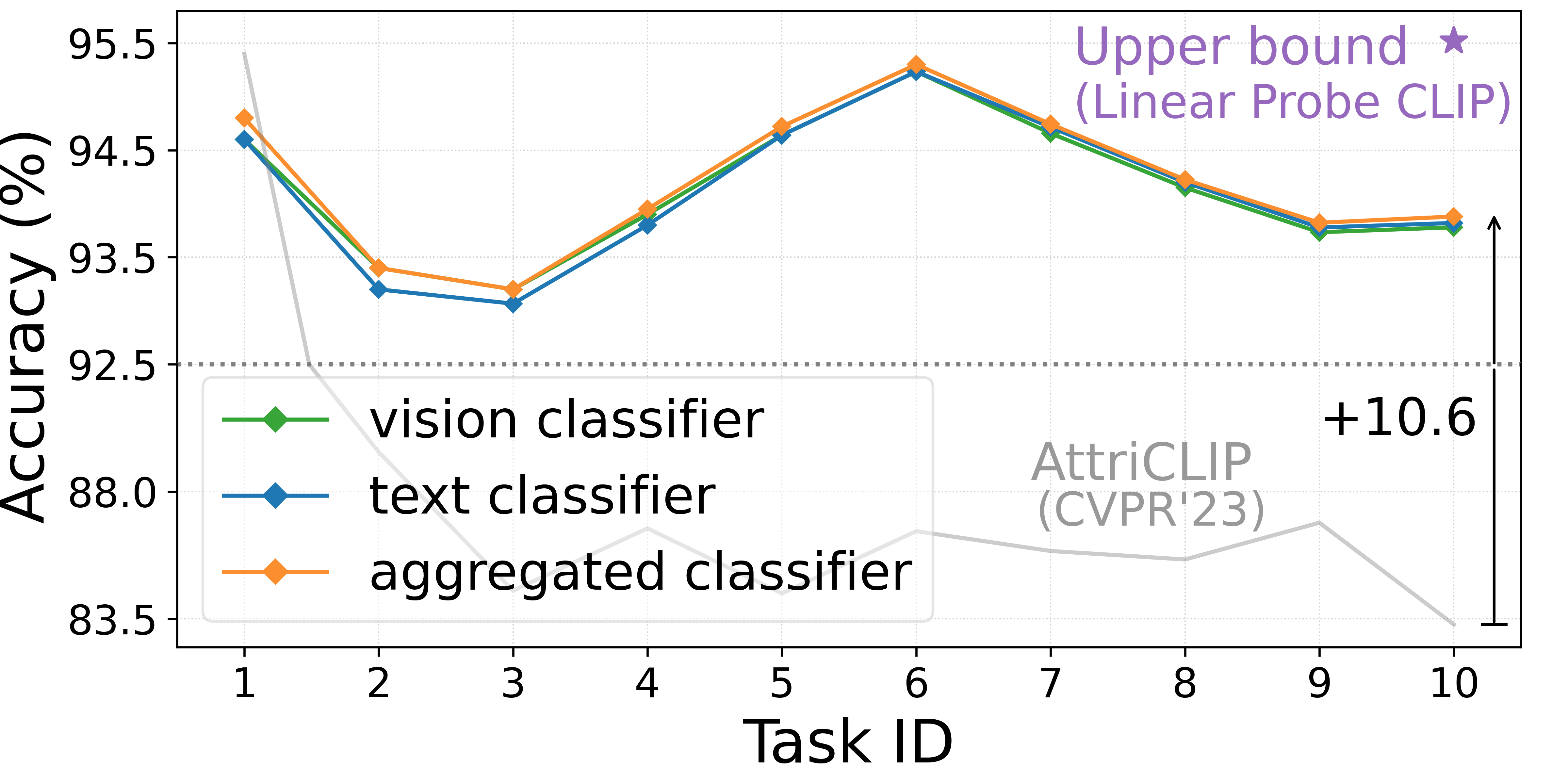}
    \caption{Accuracy evolution of ProTPS's different classifiers on ImageNet100~\cite{deng2009imagenet} under CI setting.}
    \label{fig:align_text_vision}
\end{figure}

\textbf{ImageNet100.} In Tab.~\ref{exp: imagenet100}, our ProTPS outperforms the previous SOTA AttriCLIP~\cite{wang2023attriclip} by +10.6\% on ImageNet100. The accuracy of ProTPS is only 1.6\% distance to the upper bound. We further report the accuracy of different classifiers in Fig.~\ref{fig:align_text_vision}. We can see that \textbf{ProTPS's ``vision classifier'' and ``text classifier'' are well aligned.} And the aggregated classifier achieves the best performance. Besides, the performance of ProTPS's ``text classifier'' is not capped by its ``vision classifier'' although text prompts selection is guided by prototypes.

\textbf{CIFAR100.} The results are reported in Tab.~\ref{exp: cifar100}, where $text^*$ denotes the selection and learning of our text prompts are guided by our vision prototypes in ProTPS. Our ProTPS achieves the state-of-the-art (SOTA) accuracy. Specifically, ProTPS without any data buffer outperforms CoOP~\cite{zhou2022learning} by +15.7\% and beats the previous best model AttriCLIP~\cite{wang2023attriclip} by +1.9\%. ProTPS is only 3.4\% distance from the upper bound, indicating the effectiveness of mitigating catastrophic forgetting.

\subsection{Cross-Datasets Continual Learning}
\label{sec: CDCL}
AttriCLIP~\cite{wang2023attriclip} proposes another continual learning scenario, i.e., combining two datasets into a long-sequence domain-shift task, where continually training a model across two datasets is used to evaluate the ability of knowledge transfer from the previous dataset to the new one.

\begin{table}[ht]
    \setlength{\tabcolsep}{1.5pt}
    \centering
    \begin{tabular}{cccc}
        \toprule
        Prompt & Method   & I2C & FT$\uparrow$ \\
        \midrule
        
        \multirow{4}{*}{$visual$}
        & \cellcolor{lgray}L2P-1~\cite{wang2022learning} & \cellcolor{lgray}45.7 & \cellcolor{lgray}-3.7 \\
        & \cellcolor{lgray}DualPrompt-1~\cite{wang2022dualprompt}  & \cellcolor{lgray}49.5 & \cellcolor{lgray}-4.5 \\
        & \cellcolor{lgray}CODA-P-1~\cite{smith2023coda}  & \cellcolor{lgray}66.9 & \cellcolor{lgray}-5.4 \\
        & \cellcolor{lgray}HiDe-Prompt-1~\cite{wang2023hide}   & \cellcolor{lgray}77.2 & \cellcolor{lgray}-3.8 \\
        \cline{1-1}
        \arrayrulecolor{gray}\cline{1-1}\arrayrulecolor{black}
        \multirow{4}{*}{$text$} &\cellcolor{gray}CoOp-1~\cite{zhou2022learning} (1000)  &\cellcolor{gray}61.1 & \cellcolor{gray}-6.5 \\
        &\cellcolor{gray}Continual-CLIP~\cite{thengane2022continualclip}  & \cellcolor{gray}73.4 &\cellcolor{gray}0 \\
        &\cellcolor{gray}Zero-Shot CLIP~\cite{radford2021learning} & \cellcolor{gray}78.3 & \cellcolor{gray}0 \\
        & \cellcolor{gray}AttriCLIP~\cite{wang2023attriclip}&  \cellcolor{gray}\underline{82.3} & \cellcolor{gray}\underline{+0.9} \\
        \midrule
        
        $text^*$ & \textbf{ProTPS} (ours)   & \textbf{84.4} \textcolor{darkgreen}{(+2.1)} & \textbf{+1.1} \\
        \bottomrule
\end{tabular}%
\caption{Last accuracy of different methods on CIFAR100 under CDC setting. The models are continually fine-tuned on CIFAR100 after being trained on ImageNet100 (I2C).}
\label{exp: I2C}
\end{table}

\textbf{ProTPS's text prompts are transferable to a new dataset.} In Tab.~\ref{exp: I2C}, the models are continually finetuned on 10 tasks of CIFAR100 after being trained on 10 tasks of ImageNet100, and finally evaluated on CIFAR100. This performance is denoted as ``I2C''. ``FT'' (forward transfer) is defined as the accuracy of ``I2C'' minus the accuracy of training on CIFAR100 only. When finetuning our ProTPS on CIFAR100, each class's text prompt is initialized from the prompts learned from ImageNet100 based on the highest cosine similarity between their paired prototypes. ``\textit{\{Method\}}-1'' and ``\textit{\{Method\}}-2'' are schemes defined by AttriCLIP~\cite{wang2023attriclip} on how to expand the classifier for a new dataset (please refer to AttriCLIP~\cite{wang2023attriclip} for more details). In Tab.~\ref{exp: I2C}, our ProTPS outperforms the previous SOTA AttriCLIP~\cite{wang2023attriclip} by +2.1\% in the context of ``I2C'' and has the highest ``FT'' score. This indicates that ProTPS learns transferable semantic features encoded in text prompts that can help it to generalize better to a new dataset.

\begin{table}[ht]
\centering
\begin{tabular}{ccc}
\toprule
Prompt & Method        &  I+C \\ \hline
\multirow{4}{*}{$visual$} & \cellcolor{lgray}L2P-2~\cite{wang2022learning}                            & \cellcolor{lgray}41.6                          \\
& \cellcolor{lgray}DualPrompt-2~\cite{wang2022dualprompt}                       & \cellcolor{lgray}45.8                         \\
&  \cellcolor{lgray}CODA-P-2~\cite{smith2023coda}& \cellcolor{lgray}62.7   \\
& \cellcolor{lgray}HiDe-Prompt-2~\cite{wang2023hide}  & \cellcolor{lgray}75.3  \\
\cline{1-1}
\arrayrulecolor{gray}\cline{1-1}\arrayrulecolor{black}
\multirow{4}{*}{$text$}  &  \cellcolor{gray}CoOp-2~\cite{zhou2022learning} (1000)                         &  \cellcolor{gray}55.4                          \\
&  \cellcolor{gray}Continual-CLIP~\cite{thengane2022continualclip}                      &  \cellcolor{gray}65.9                          \\
&  \cellcolor{gray}Zero-Shot CLIP~\cite{radford2021learning}  &  \cellcolor{gray}68.7 \\
& \cellcolor{gray}AttriCLIP~\cite{wang2023attriclip}        &  \cellcolor{gray}\underline{78.3}                 \\
\midrule

$text^*$
& \textbf{ProTPS} (ours) & \textbf{85.6} \textcolor{darkgreen}{(+7.3)} \\

\bottomrule
\end{tabular}
\caption{Last accuracy of different methods when continually trained on ImageNet100 and CIFAR100 (I+C) in sequence under CDC setting.}
\label{exp: I+C}
\end{table}

\textbf{ProTPS is suitable for cross-datasets continual learning.} In Tab.~\ref{exp: I+C}, models are evaluated on both datasets. For ProTPS, when start fine-tuning on CIFAR100, we expand our $classifier_1$ for CIFAR100 classes, and each class's text prompt is still initialized from the prompts learned from ImageNet100 based on the same rule. ProTPS beats visual-prompt-based and text-prompt-based methods, outperforming the AttriCLIP~\cite{wang2023attriclip} by +7.3\%.

\begin{table*}[h]
    \centering
    \begin{tabular}{cccccccc}
        \toprule
        Dataset & Authentic CL & Setting & Distribution & \#Classes & Images  &Fine-grained  \\
        \midrule
        Rotation MNIST~\cite{lopez2017gradient} & $\times$ & DI & balanced & 10 & synthetic & $\times$ \\
        Permutation MNIST~\cite{kirkpatrick2017overcoming} & $\times$ & DI & balanced &10 & synthetic & $\times$\\
        NS-MiniImageNet~\cite{mai2022online} & $\times$ & DI & balanced & 100 & synthetic  & $\times$\\
        CLEAR~\cite{lin2021clear} & $\times$ & DI & balanced & 100 & real & $\times$\\
        DS-ImageNet-R~\cite{smith2023coda}   & $\times$ & CDI\tablefootnote{DS-ImageNet-R introduces domain shifts that occur between classes, rather than within each class, as reflected in our training data.}  &  balanced  & 200 & real & $\times$ \\
        CORe50-NIC~\cite{lomonaco2017core50}  & $\times$ & CDI\tablefootnote{CORe50-NIC aims to evaluate the model's generalization to unseen domains, rather than its performance on domains learned incrementally, as demonstrated by our test protocol.}   &  balanced  & 50 & real & $\times$\\
        \midrule
        \textbf{Marine112} (Ours)  & \checkmark  & CDI &  longtail  &  112 & real  & \checkmark \\
        \bottomrule
    \end{tabular}
    \caption{Comparison of Marine112 with public datasets with domain shift. ``CL'': Continual learning. ``DI'': Domain incremental learning. ``CDI'': Class and domain incremental learning.}
    \label{table: marine112 comparison}
\end{table*}

\begin{figure}[h]
    \centering
    \includegraphics[width=0.45\textwidth]{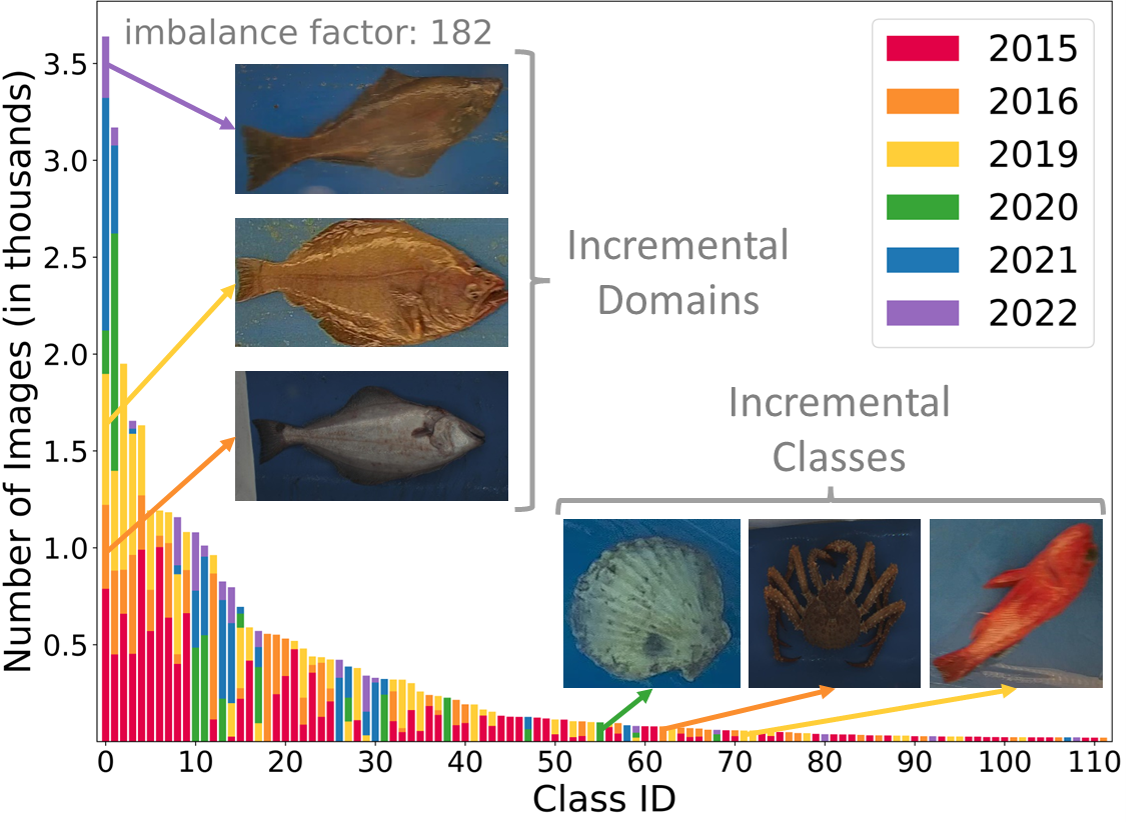}
    \captionof{figure}{\textbf{Marine112} captures both ecological dynamics (incremental domains and classes) and natural longtail distribution.}
    \label{fig:fish_stack_histogram}
\end{figure}

\begin{table*}[h]
    \centering
    \begin{tabular}{cccccccc}
        \toprule
         Prompt &  Method & 2015 & 2016 & 2019 & 2020 & 2021 & 2022 \\
        \hline
         \multirow{2}{*}{$visual$} & \cellcolor{lgray}CODA-P~\cite{smith2023coda} & \cellcolor{lgray}71.2 & \cellcolor{lgray}42.1  & \cellcolor{lgray}33.6 & \cellcolor{lgray}13.4 & \cellcolor{lgray}12.7 & \cellcolor{lgray}13.2  \\
          & \cellcolor{lgray}HiDe-Prompt~\cite{wang2023hide} & \cellcolor{lgray}73.4 & \cellcolor{lgray}\underline{50.8}  & \cellcolor{lgray}\underline{45.2} & \cellcolor{lgray}\underline{22.2} & \cellcolor{lgray}\underline{22.3} & \cellcolor{lgray}\underline{22.5}  \\
          \cline{1-1}
          \arrayrulecolor{gray}\cline{1-1}\arrayrulecolor{black}
         \multirow{4}{*}{$text$} & \cellcolor{gray}Continual-CLIP~\cite{thengane2022continualclip} & \cellcolor{gray}12.7 & \cellcolor{gray}11.8  & \cellcolor{gray}10.2 &\cellcolor{gray}9.6 & \cellcolor{gray}10.1 & \cellcolor{gray}9.5  \\
          & \cellcolor{gray}Zero-Shot CLIP~\cite{radford2021learning} & \cellcolor{gray}15.4 & \cellcolor{gray}13.1  & \cellcolor{gray}11.6 &\cellcolor{gray}10.3 & \cellcolor{gray}10.5 &\cellcolor{gray}9.6   \\
          & \cellcolor{gray}CoOp~\cite{zhou2022learning} (1000) & \cellcolor{gray}65.3 & \cellcolor{gray}30.2 & \cellcolor{gray}15.1 & \cellcolor{gray}10.2 & \cellcolor{gray}11.4 &\cellcolor{gray}10.3\\
          &  \cellcolor{gray}AttriCLIP~\cite{wang2023attriclip} & \cellcolor{gray}\underline{76.1} & \cellcolor{gray}48.3  & \cellcolor{gray}34.8 & \cellcolor{gray}11.9 & \cellcolor{gray}13.6 & \cellcolor{gray}13.9  \\
          \midrule
          $text^*$ & \textbf{ProTPS} (ours)  & \textbf{76.6} & \textbf{70.5}  & \textbf{61.5} & \textbf{59.7} & \textbf{58.1} & \textbf{55.6} \textcolor{darkgreen}{(+33.1)}\\
        \midrule
         &  Upper bound & - & -  & - & - & - & 78.2 \\
        \bottomrule
    \end{tabular}
    \caption{Accuracy on real-world Marine112 dataset under CDI setting.}
    \label{exp: marine112}
\end{table*}

\subsection{Class-and-Domain Incremental Learning}
\label{sec: CDI}

\textbf{Marine112} consists of 112 marine species collected over a span of \textbf{six years}. Each year, fishing ships equipped with cameras were deployed to specific oceanic regions to capture images of marine species, aiming to analyze changes in the marine ecosystem. Over time, the dataset reflects the emergence of new species, such as crabs and shellfish, alongside notable \textbf{domain shifts} in captured species, as illustrated in Fig.~\ref{fig:fish_stack_histogram}. These shifts arise due to variations in illumination conditions, differences in camera brands, and resolution disparities between years. Notably, the appearance of the same species across different years may vary significantly, with perceptible differences in color and texture, as shown in Fig.~\ref{fig:fish_stack_histogram}. This evolving dataset captures both ecological dynamics and the technical challenges associated with \textbf{longtail}~\cite{liu2019large} distribution observed in nature. Tab.~\ref{table: marine112 comparison} contrasts our real-world Marine112 dataset with several public datasets commonly used for continual learning. Marine112’s unique distinction lies in its \underline{authentic}, real-world data, making it intrinsically well-suited for tackling the class and domain incremental (CDI) learning task. Further details, such as the test set split, are in Appendix~\ref{sec: marine112}.

\textbf{ProTPS learns more unique text prompts than existing methods.} In Tab.~\ref{exp: marine112}, ProTPS sets the strongest baseline on Marine112. For visual-prompt-based methods, we only benchmarked CODA-P~\cite{smith2023coda} and HiDe-Prompt~\cite{wang2023hide} because they achieved better performance than others in previous experiments.


\begin{table}
    \setlength{\tabcolsep}{3pt}
    \centering
    \begin{tabular}{ccccc}
        \toprule
          PL & Prompt-Selection & STC & $Loss_{pp} $&  Last acc.  \\
        \hline
          refined & weight-top-1  & \checkmark & \checkmark  & \textbf{83.28}\\
          \hline
          refined  & weight-top-1 & $\times$ & \checkmark  & 80.95 \textcolor{lightred}{(-2.33)}\\
          refined & weight-top-1  & \checkmark & $\times$  & 80.87 \textcolor{lightred}{(-2.41)}\\
          frozen  & weight-top-1 & \checkmark & \checkmark  & 80.65 \textcolor{lightred}{(-2.63)}\\
        refined  & $\times$  & \checkmark & \checkmark  & 77.62 \textcolor{lightred}{(-5.66)}\\
          scratch  & weight-top-1  & \checkmark & \checkmark  & 58.42 \textcolor{lightred}{(-24.86)}\\

        \bottomrule
    \end{tabular}
    \caption{Ablation study on CIFAR100.``PL'': Prototype Learning strategy. ``STC'': Sampled Text Classifier strategy.}
    \label{exp: ablation on cifar100}
\end{table}


\begin{figure}[h]
    \centering
    \includegraphics[width=0.48\textwidth]{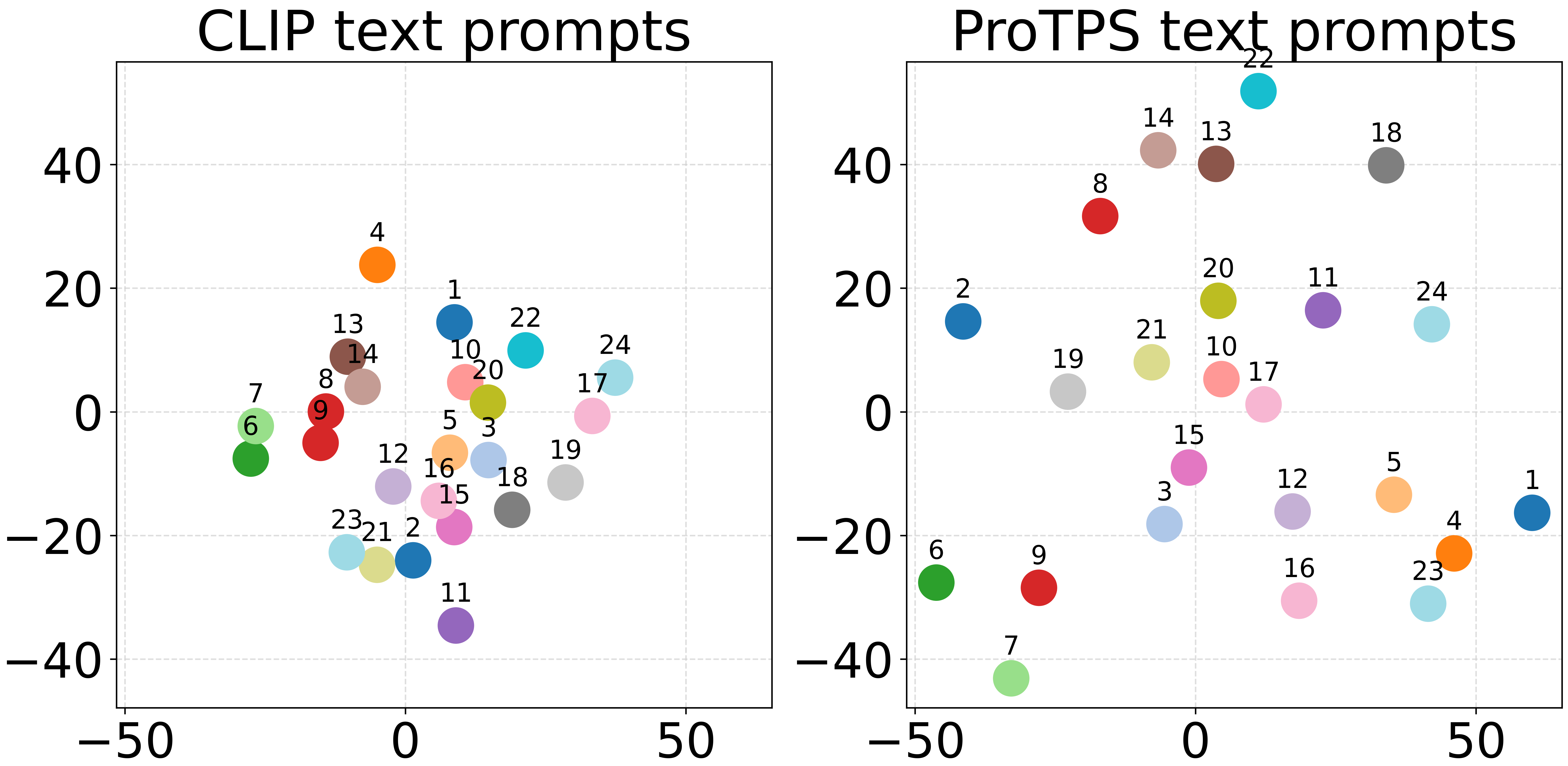}
    \captionof{figure}{Comparison of t-SNE~\cite{van2008visualizing} visualization on text prompts for 24 dog breeds from ImageNet100~\cite{deng2009imagenet}.}
    \label{fig:tsne_text_prompts}
\end{figure}

\subsection{Ablation Study}
\label{sec: ablation}

Tab.~\ref{exp: ablation on cifar100} shows the impact of each proposed module in ProTPS. The proposed prototype learning strategy is denoted as ``refined''. We can see it lays the foundation for the success of ProTPS's text prompt learning guided by learned prototypes. It outperforms learning prototypes ``from scratch'' or freezing them after the proposed initialization method, referred to as ``frozen''. The proposed text prompt selection method, denoted as ``weight-top-1'', also brings a \textbf{significant improvement}. Both proposed $sampled$ $classifier_2$ strategy and ``prompt-prototype contrastive loss'' also bring non-negligible improvement.

\textbf{t-SNE visualization of text prompts.} In Fig.~\ref{fig:tsne_text_prompts}, we use t-SNE~\cite{van2008visualizing} to display the distribution of text prompts in the feature space for 24 dog breeds from ImageNet100. This is done after the text prompts are processed by the text encoder. We can see our text prompts are more distinct in the feature space than CLIP's text prompt templates.


\begin{figure}[h]
    \centering
    \includegraphics[width=0.48\textwidth]{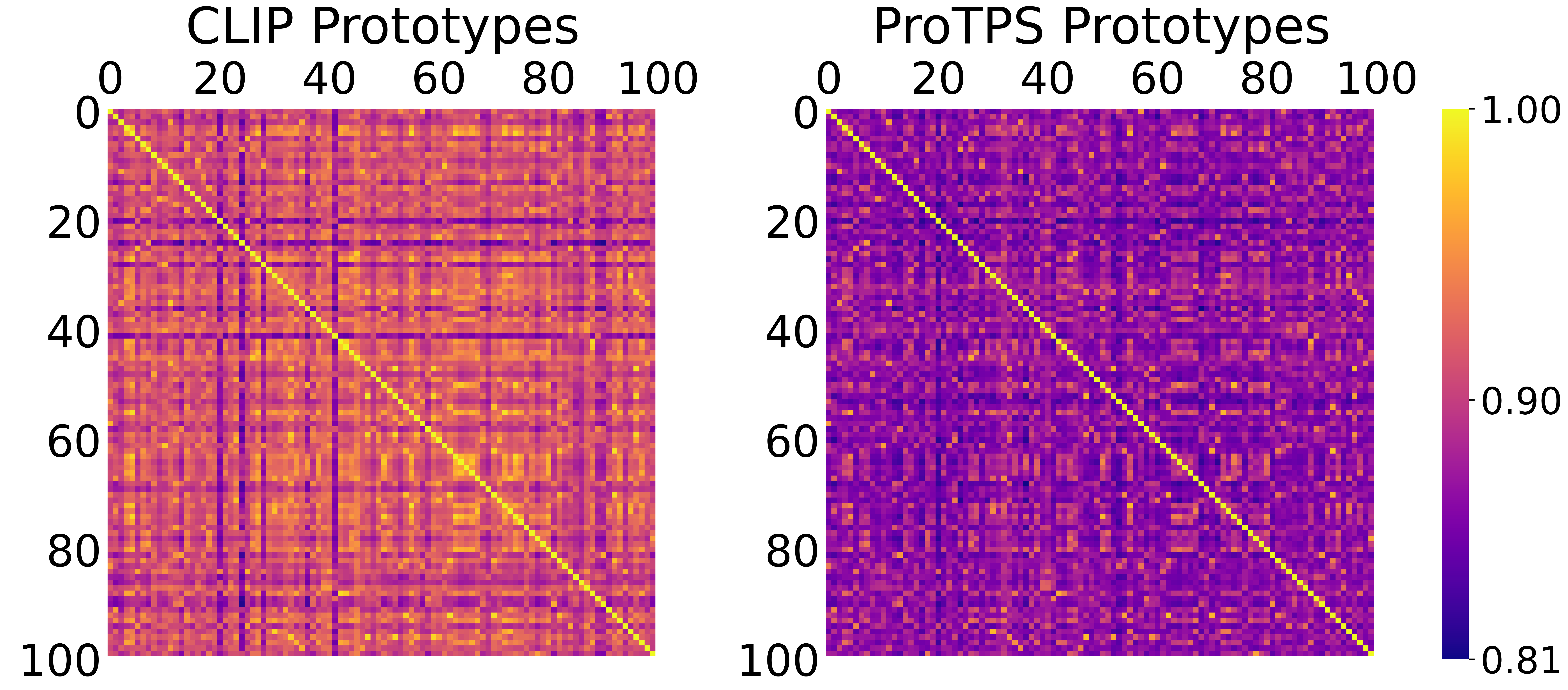}
    \captionof{figure}{Similarity matrix of prototypes on CIFAR100. Left: Initialized prototypes via the frozen CLIP image encoder. Right: ProTPS's refined vision prototypes.}
    \label{fig:vision proto matrix}
\end{figure}


\textbf{Visualization of vision prototypes.} To further confirm the improvement of our refined vision prototypes compared to the initialization at the beginning of each task, we visualize the cosine similarity matrix among prototypes in Fig.~\ref{fig:vision proto matrix}. Despite freezing earlier learned prototypes in later tasks, ProTPS's vision prototypes become more distinguishable after 10 learning tasks on CIFAR100 with our training recipe, as indicated by the darker color outside the diagonal.

\textbf{Visualization of text prompts on images.} To verify that our learned unique text prompts can attend to region-level image details even among similar classes, we visualize the image contents corresponding to each class's text prompts in Fig.~\ref{fig: prompt vis} on ImageNet100. The ``CLIP'' row shows the attention of the original CLIP's ensembled prompts on paired images. Details and more visualizations are in Appendix~\ref{sec: CLIP templates}. We can see that our ProTPS's learned text prompts can focus on specific regions such as face, body, tail, beak, etc.

%% file: sec/5_conclu.tex
\section{Conclusion}
\label{sec:conclu}
We propose ProTPS, which continually learns unique text prompts to mitigate ``catastrophic forgetting''. ProTPS leverages global image-level vision prototypes to guide the selection and learning of regional semantic features encoded in text prompts. Our experiments demonstrate superior performance over existing state-of-the-art methods in CI, CDC, and CDI learning settings. Additionally, we introduced Marine112, a real-world dataset containing 112 marine species over a span of 6 years and authentically suitable for CDI setting, which brings new challenges to the community. We believe our ProTPS and Marine112 can pave the way for more practical continual learning.


%% file: sec/X_suppl.tex
\clearpage
\setcounter{page}{1}



\appendix
\renewcommand{\thesection}{\Alph{section}} 
\renewcommand{\thesubsection}{\thesection.\arabic{subsection}} 
\renewcommand{\thesubsubsection}{\thesubsection.\arabic{subsubsection}} 


\section{Classes of ImageNet100}
\label{sec: Classes of ImageNet100}

Following AttriCLIP~\cite{wang2023attriclip}, we use the same 100 classes from ImageNet~\cite{deng2009imagenet} to build ImageNet100. The class names are: \textit{`Robin', `Gila monster', `hognose snake', `garter snake', `green mamba', `garden spider', `lorikeet', `goose', `rock crab', `fiddler crab', `American lobster', `little blue heron', `American coot', `Chihuahua', `Shih Tzu', `Papillon', `toy terrier', `Walker hound', `English foxhound', `borzoi', `Saluki', `American Staffordshire Terrier', `Chesapeake Bay Retriever', `Vizsla', `Kuvasz', `Komondor', `Rottweiler', `Doberman', `Boxer', `Great Dane', `Standard Poodle', `Mexican hairless', `coyote', `African hunting dog', `red fox', `tabby', `meerkat', `dung beetle', `walking stick', `leafhopper', `hare', `wild boar', `gibbon', `langur', `ambulance', `bannister', `bassinet', `boathouse', `bonnet', `bottlecap', `car wheel', `chime', `cinema', `cocktail shaker', `computer keyboard', `Dutch oven', `football helmet', `gasmask', `hard disc', `harmonica', `honeycomb', `iron', `jean', `lampshade', `laptop', `milk can', `mixing bowl', `modem', `moped', `shower cap', `mousetrap', `obelisk', `park bench', `pedestal', `pickup', `pirate', `purse', `reel', `rocking chair', `rotisserie', `safety pin', `sarong', `ski mask', `slide rule', `stretcher', `theater curtain', `throne', `tile roof', `tripod', `tub', `vacuum', `window screen', `wing', `head cabbage', `cauliflower', `pineapple', `carbonara', `chocolate sauce', `gyromitra', `mushroom'}

\section{Marine112 Dataset}
\label{sec: marine112}

We believe our Marine112 dataset is of great importance to the continual learning community, for the following reasons: (1) it features overlapped marine classes among different tasks with varied imaging qualities, i.e., domain shift; (2) it reflects the longtail~\cite{liu2019large} (imbalanced) distribution observed in nature; and (3) it is naturally suited for the class and domain incremental (CDI) learning. In Marine112, there are 112 classes across 6 different years as shown in Fig.~\ref{fig:fish_stack_histogram}. This dataset is inspired by existing work~\cite{Kiefer_2024_WACV,yang2024sea,mei2021video,mei2021absolute,zheng2023progressive,kiefer20232nd,mei2022hcil,zhang2021u3d,wang2021hvps,mei2022unsupervised,mei2024esa,mei2024continual}
 


\subsection{Species Names}
The labeled class names are: \textit{`arrowtooth flounder', `rex sole', `pacific ocean perch', `flathead sole', `walleye pollock', `shortspine thornyhead', `pacific halibut', `northern rockfish', `sablefish', `pacific cod', `petrale sole', `starfish', `northern rocksole', `dungeness crab', `english sole', `dover sole', `southern rock sole', `slender sole', `kamchatka flounder', `atka mackerel', `yellow irish lord', `eulachon', `blackspotted rockfish', `giant grenadier', `darkfin sculpin', `dusky rockfish', `big skate', `spotted ratfish', `rougheye rockfish', `pacific sanddab', `sea anemone', `longnose skate', `shortraker rockfish', `harlequin rockfish', `butter sole', `redstripe rockfish', `strongylocentrotus', `sturgeon poacher', `greenstriped rockfish', `berryteuthis magister', `searcher', `sharpchin rockfish', `yellowfin sole', `pacific tomcod', `podothecus accipenserinus', `fusitriton oregonensis', `metridium farcimen', `bathyraja unidentified', `actinauge verrillii', `bairidi tanner crab', `pacific herring', `chum salmon', `red banded rockfish', `spectacled sculpin', `silvergray rockfish', `jellyfish', `great sculpin', `prowfish', `scallop', `lingcod', `ctenodiscus crispatus', `pandalopsis dispar', `golden king crab', `popeye grenadier', `gorgonocephalus eucnemis', `scissortail sculpin', `armorhead sculpin', `paragorgia arborea', `darkblotched rockfish', `chlamys', `pacific octopus', `longspine thornyhead', `rosethorn rockfish', `starry flounder', `spinyhead sculpin', `kelp greenling', `mud skate', `sawback poacher', `yellowtail rockfish', `pandalus', `shrimp', `alaska plaice', `dark rockfish', `bathymaster signatus', `ebony eelpout', `cyanea capillata', `toad lumpsucker', `pycnopodia helianthoides', `northern lampfish', `pacific sand fish', `solaster', `cucumaria fallax', `eelpout unidentified', `chrysaora melanaster', `shortfin eelpout', `spot shrimp', `ceramaster', `bigmouth sculpin', `capelin', `rock greenling', `stegophiura ponderosa', `leopard skate', `yelloweye rockfish', `parastichopus californicus', `allocentrotus fragilis', `hippasteria', `black rockfish', `sea pen', `skate egg case', `serpula', `crossaster papposus', `white blotched skate'}

\begin{figure}[!h]
    \centering
    \includegraphics[width=0.98\columnwidth]{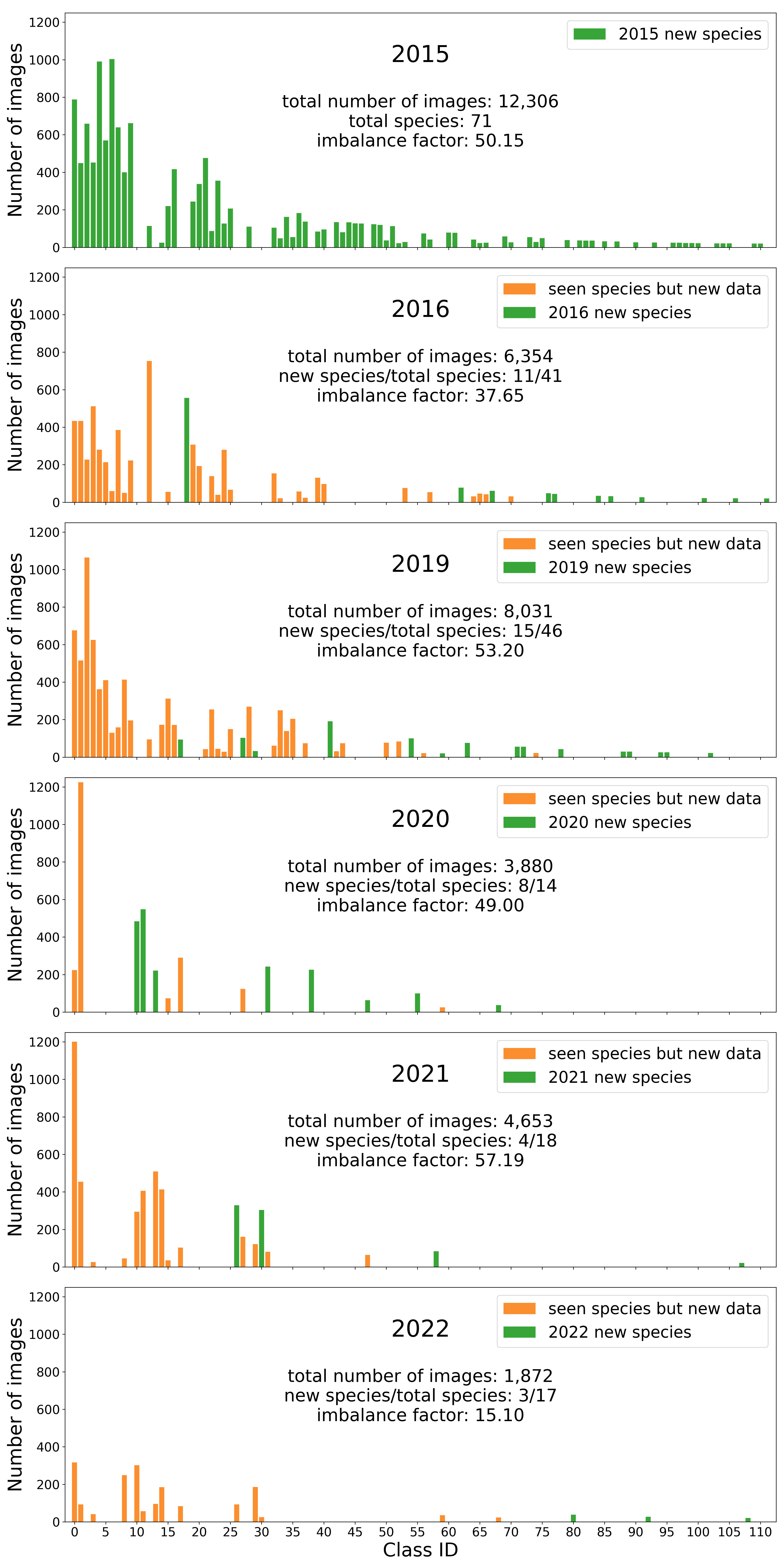}
    \caption{\textbf{Marine112.} Data distribution in each year.}
    \label{fig:fish_hist_total_year}
\end{figure}

\subsection{Data Distribution}
In Marine112, annual data collection includes new data for both newly discovered and previously identified classes. The latter introduces domain shifts due to illumination, pose, imaging qualities of different types of cameras, etc, as shown in Fig.~\ref{fig:fish_stack_histogram}. Detailed data distribution of each year is in Fig.~\ref{fig:fish_hist_total_year}. The imbalance factor, calculated as the ratio of the most common class's instance count to that of the least common class, indicates the degree of class imbalance.

\subsection{Testing Data Split}
\begin{figure}[htbp]
    \centering
    \includegraphics[width=0.48\textwidth]{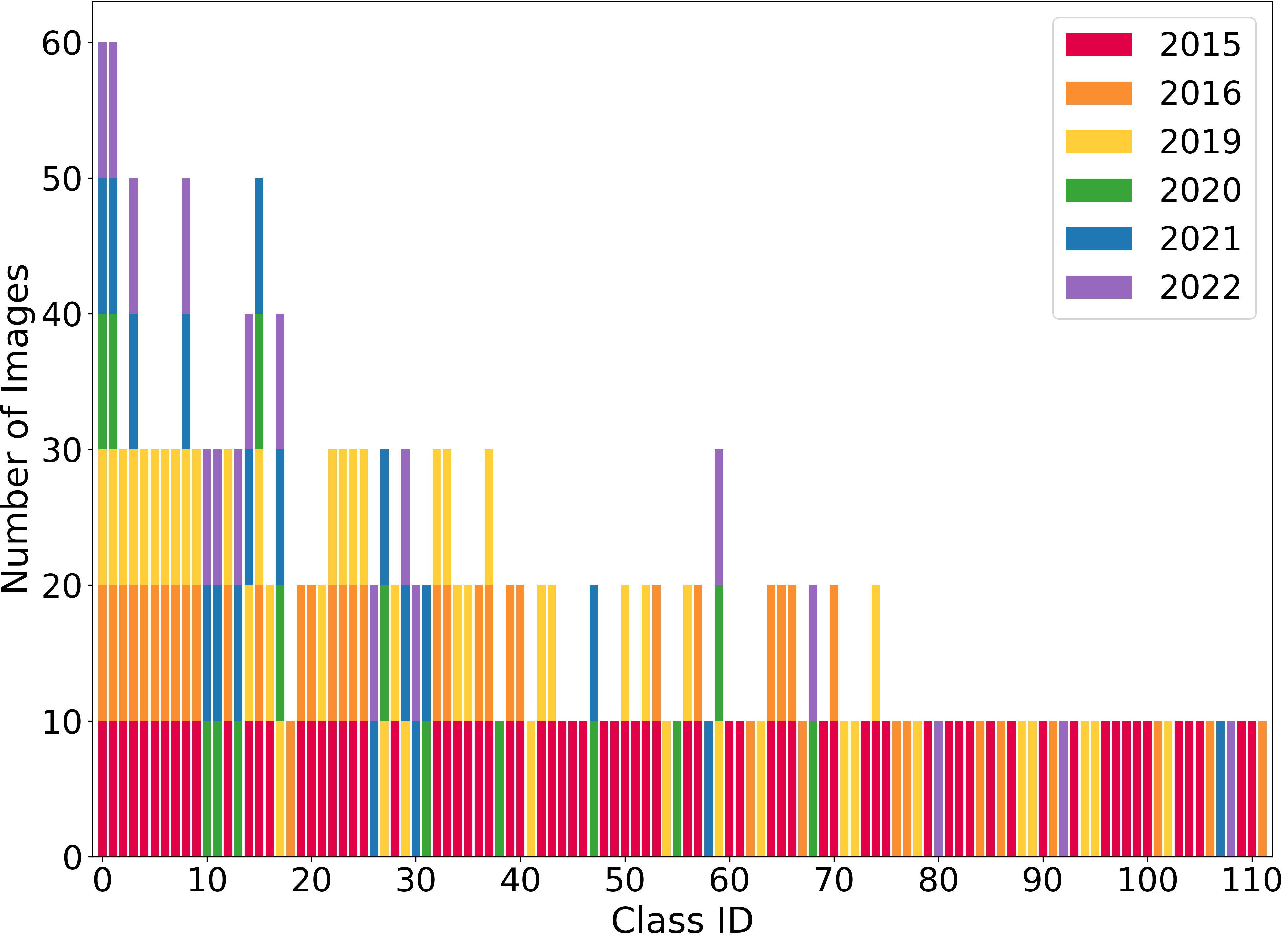}
    \caption{\textbf{Marine112.} Number of kept test data from each year.}
    \label{fig:test_data_distribution_dpi300}
\end{figure}
As mentioned in the previous section, new data of previously seen classes usually bring domain shift. Therefore, each year, no matter the new classes or old classes, we always randomly hold 10 new images from each collected class as testing data. And the accuracy of each year $t, t \in \{2015, 2016, \ldots, 2022\}$ is evaluated on testing images from all the trained years so far (i.e., year 2015, ..., t). Fig.~\ref{fig:test_data_distribution_dpi300} shows the number of kept test data from each year.

\textbf{Justification.} There is a real-world application scenario for the above test data split protocol. Initially, in 2015, thousands of images were collected via electronic monitoring from different commercial fishing boats to ensure that fish populations in targeted areas were sustainably managed. The identification of marine species from images is an expensive time-consuming process that requires the consulting of fishery specialists. Therefore, deep learning models serve as valuable tools to efficiently process such data. To train a deep learning model, we ask fishery specialists to label a portion of the collected data. Our model then is expected to identify remaining unlabeled images, reducing the labor cost required to gain an accurate understanding of the species and numbers caught. Our goal is to train a model continually with a subset of each year’s data, manually labeled, and apply it to unlabelled data of previous years and the current year. Thus, in our testing protocol, once we train the model on the 2015 labeled training data, we test it on the held 2015 testing data. Subsequently, we continue to train the model on 2016 training data, and then test the model on both 2015 and 2016 testing data. Therefore, our testing split protocol can reflect the performance of the continually trained model on \textbf{accumulated} unidentified data from incrementally exposed domains and classes. 




\section{Implementation Details} 
\label{sec: training details}

\subsection{Training Details}
\label{sec: ProTPS training details}
\begin{figure}[htbp]
    \centering
    \includegraphics[width=0.98\columnwidth]{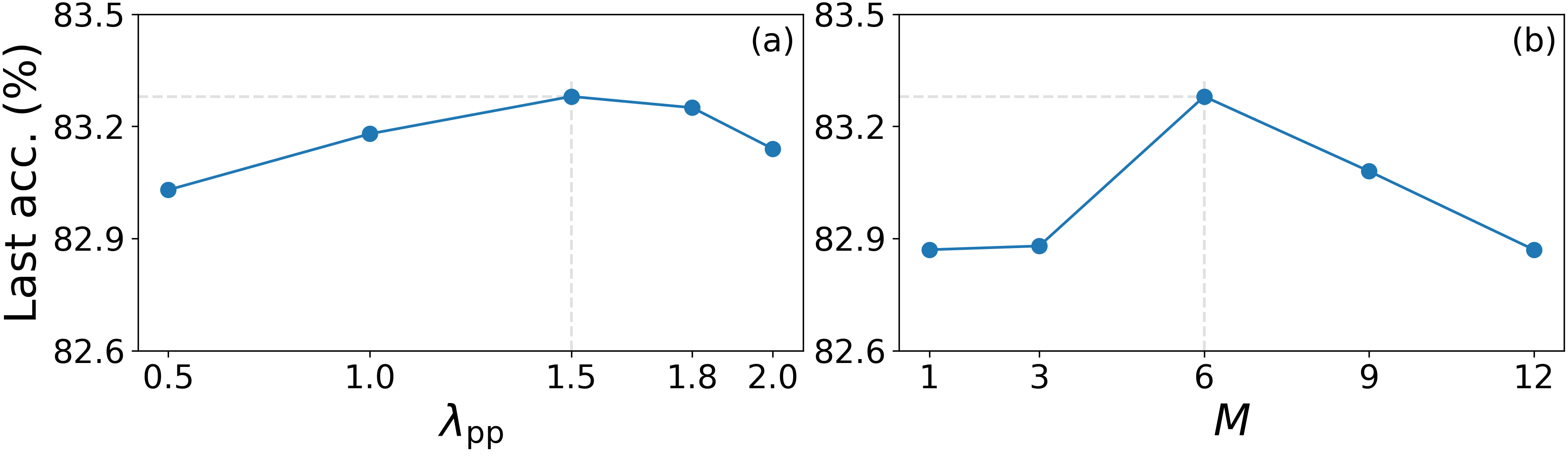}
    \caption{Comparison of different (a) loss factor $\lambda_{pp}$ values and different (b) prompt length $M$ values on CIFAR100~\cite{krizhevsky2009learning}.}
    \label{fig:hyper search}
\end{figure}

We train our ProTPS for 10 epochs on each task for all datasets. SGD is adopted as the optimizer with an initial learning rate of 0.001 and follows a cosine decay schedule to a final learning rate of 1e-4. The weight decay is 0, and the batch size is 128 on two V100 GPUs. Each text prompt $P_i$ is a learnable embedding with the dimension of [$M$, 768] where $M$ is the prompt length and 768 is the feature dimension output from the last projection layer of CLIP's image encoder. We observe that the last accuracy is not sensitive to the loss factor $\lambda_{pp}$ and prompt length $M$ indicated in Fig.~\ref{fig:hyper search} thus we use $\lambda_{pp}$=1.5 and $M$ = 6 for our ProTPS. The average results over 3 runs are reported for all methods in the paper.

\textbf{Other baselines details.} For baseline methods, we use their own official released code and follow their own training recipes. CoOp~\cite{zhou2022learning} is the only method in all tables storing 1000 data samples during training thus denoted as ``CoOp~\cite{zhou2022learning} (1000)'' in all tables. The prompts of CoOp~\cite{zhou2022learning} and AttriCLIP~\cite{wang2023attriclip} are continually trained in a sequence of tasks. For ``Zero-Shot CLIP'', ``Prototypes-CLIP'', and ``Linear Probe CLIP'', we implement them based on the official CLIP's open-source code. All reported CLIP-based methods use the same original CLIP pre-trained transformer~\cite{vaswani2017attention} as the text encoder. All reported methods use the original CLIP pre-trained ViT-L/14~\cite{dosovitskiy2020image} as the image encoder. All implementation codes and the Marine112 dataset will be released.

\textbf{Why use CLIP pre-trained image encoder for all reported methods?} Firstly, our ProTPS is based on CLIP architecture similar to existing text-prompt-based methods~\cite{zhou2022learning, thengane2022continualclip, wang2022s, wang2023attriclip} which use frozen CLIP pre-trained image encoder and text encoder. Secondly, as mentioned in DualPrompt~\cite{wang2022dualprompt}, when visual-prompt-based methods~\cite{wang2022dualprompt, wang2022learning, smith2023coda, wang2023hide} use the image encoder pre-trained on ImageNet, they struggle with fairness when evaluated on the ImageNet-related benchmarks such as ImageNet100 and ImageNet-R~\cite{wang2022dualprompt}. Thus we use CLIP pre-trained ViT-L/14~\cite{dosovitskiy2020image} as the image encoder for all methods.

\textbf{Classifiers aggregation methods.} Tab.~\ref{exp: unite method} shows aggregating ProTPS's dual classifiers via the proposed average operation is better than the max operation.

\begin{table}[t]
    \centering
    \begin{tabular}{cc}
        \toprule
        Aggregation   & Last acc. \\
        \midrule
        Average & \textbf{93.92}\\
        Max &  93.86 \textcolor{lightred}{(-0.06)} \\
        \bottomrule
    \end{tabular}
    \captionof{table}{Different aggregation methods on ImageNet100.}
    \label{exp: unite method}
\end{table}

\subsection{Evaluation Metric Details}
\label{sec: metric}
As emphasized by previous works~\cite{wang2022dualprompt, smith2023coda}, ``Average Accuracy'' is the overall evaluation metric for continual learning, which includes two aspects: greater learning capacity and less catastrophic forgetting. 

However, real-world continual learning datasets may be imbalanced, \eg Marine112. Thus, we define the \textit{generalized} average accuracy to evaluate continual learning models. Consider a series of $T$ tasks $D=
\{D_1, . . . , D_T\}$. Each task has its own training data, $D_{t, train} =\{(x_{t, train}^{i}, y_{t, train}^{i})\}_{i=1}^{n_{t, train}}$ comprising $n_{t, train}$ instances $x_{t, train}^{i}$ and their respective labels $y_{t, train}^{i}$, and testing data, $D_{t, test} =\{(x_{t, test}^{i}, y_{t, test}^{i})\}_{i=1}^{n_{t, test}}$ comprising $n_{t, test}$ instances $x_{t, test}^{i}$ and their respective labels $y_{t, test}^{i}$. After the model finishes training on the task $t$, we compute \textit{generalized} average accuracy $A_t$ as follows:
\begin{equation}
A_t = \frac{ \sum_{m=1}^{t} \sum_{i=1}^{n_{m, test}} equal(model(x_{m,test}^i), y_{m, test}^{i})}{ \sum_{m=1}^{t} n_{m, test}},
\end{equation}
where the numerator is the number of true predictions on testing data from all exposed tasks (i.e., task $1 ,2 ...t$) and the denominator is the total number of involved testing data. When testing data from each task is balanced, $A_t$ is exactly the widely-used ``Average Accuracy'' well-defined in previous works such as~\cite{wang2022dualprompt} (see its Appendix C). Similarly, we refer to the final task's $A_T$ as the ``Last Accuracy''.

\subsection{More Results on Class Incremental Learning}
\label{sec: SLCA+APG}


\begin{table*}[h]
\small
\centering
\begin{tabular}{cccc}
\hline
Prompt & Method        &  CIFAR100 & ImageNet100\\ \hline
- & SLCA~\cite{zhang2023slca} & 79.1 & 56.3 \\
\cline{1-1}
\arrayrulecolor{gray}\cline{1-1}\arrayrulecolor{black}
\multirow{5}{*}{$visual$} & \cellcolor{lgray}L2P~\cite{wang2022learning}                            & \cellcolor{lgray}49.4 &  \cellcolor{lgray}48.3\\
& \cellcolor{lgray}DualPrompt~\cite{wang2022dualprompt}                       & \cellcolor{lgray}54.0 & \cellcolor{lgray}52.6                        \\
& \cellcolor{lgray}APG~\cite{tang2023prompt}  & \cellcolor{lgray}54.2 &  \cellcolor{lgray}53.1  \\
&  \cellcolor{lgray}CODA-P~\cite{smith2023coda}& \cellcolor{lgray}72.3 & \cellcolor{lgray}75.2   \\
& \cellcolor{lgray}HiDe-Prompt~\cite{wang2023hide}  & \cellcolor{lgray}81.0 &  \cellcolor{lgray}82.9  \\
\cline{1-1}
\arrayrulecolor{gray}\cline{1-1}\arrayrulecolor{black}
\multirow{4}{*}{$text$}  &  \cellcolor{gray}CoOp~\cite{zhou2022learning} (1000)                         &  \cellcolor{gray}67.6 & \cellcolor{gray}79.3                          \\
&  \cellcolor{gray}Continual-CLIP~\cite{thengane2022continualclip}                      &  \cellcolor{gray}73.4 & \cellcolor{gray}75.4                          \\
&  \cellcolor{gray}Zero-Shot CLIP~\cite{radford2021learning}  &  \cellcolor{gray}78.3& \cellcolor{gray}75.1 \\
& \cellcolor{gray}AttriCLIP~\cite{wang2023attriclip}        &  \cellcolor{gray}\underline{81.4} & \cellcolor{gray}\underline{83.3}                 \\
\midrule

$text^*$ & \textbf{ProTPS} (ours) & \textbf{83.3} \textcolor{darkgreen}{(+1.9)} & \textbf{93.9} \textcolor{darkgreen}{(+10.6)} \\

\bottomrule
\end{tabular}
\caption{Last accuracy on CIFAR100~\cite{krizhevsky2009learning} and ImageNet100~\cite{deng2009imagenet} under CI setting.}
\label{exp: SLCA & APG}
\end{table*}

\begin{table*}[ht]
    \small
    \centering
    \begin{tabular}{ccccc}
        \toprule
        Prompt & Method  & C & I2C & FT$\uparrow$ \\
        \midrule
        
        $v+t$  & S-liPrompts~\cite{wang2022s} & 58.9 & 53.3 & -5.6 \\
        \cline{1-1}
        \arrayrulecolor{gray}\cline{1-1}\arrayrulecolor{black}
        \multirow{4}{*}{$visual$}
        & \cellcolor{lgray}L2P-1~\cite{wang2022learning}& \cellcolor{lgray}49.4 & \cellcolor{lgray}45.7 & \cellcolor{lgray}-3.7 \\
        & \cellcolor{lgray}DualPrompt-1~\cite{wang2022dualprompt} & \cellcolor{lgray}54.0 & \cellcolor{lgray}49.5 & \cellcolor{lgray}-4.5 \\
        & \cellcolor{lgray}CODA-P-1~\cite{smith2023coda} & \cellcolor{lgray}72.3 & \cellcolor{lgray}66.9 & \cellcolor{lgray}-5.4 \\
        & \cellcolor{lgray}HiDe-Prompt-1~\cite{wang2023hide}  & \cellcolor{lgray}81.0 & \cellcolor{lgray}77.2 & \cellcolor{lgray}-3.8 \\
        \cline{1-1}
        \arrayrulecolor{gray}\cline{1-1}\arrayrulecolor{black}
        \multirow{4}{*}{$text$} &\cellcolor{gray}CoOp-1~\cite{zhou2022learning} (1000) &\cellcolor{gray}67.6 &\cellcolor{gray}61.1 & \cellcolor{gray}-6.5 \\
        &\cellcolor{gray}Continual-CLIP~\cite{thengane2022continualclip}  & \cellcolor{gray}73.4 & \cellcolor{gray}73.4 &\cellcolor{gray}0 \\
        &\cellcolor{gray}Zero-Shot CLIP~\cite{radford2021learning}  & \cellcolor{gray}78.3 & \cellcolor{gray}78.3 & \cellcolor{gray}0 \\
        & \cellcolor{gray}AttriCLIP~\cite{wang2023attriclip}& \cellcolor{gray}\underline{81.4} & \cellcolor{gray}\underline{82.3} & \cellcolor{gray}\underline{+0.9} \\
        \midrule
        
        $text^*$ & \textbf{ProTPS} (ours)  & \textbf{83.3} & \textbf{84.4} \textcolor{darkgreen}{(+2.1)} & \textbf{+1.1} \\
        \bottomrule
\end{tabular}%
\caption{Last accuracy of different methods on CIFAR100~\cite{krizhevsky2009learning} under CDC setting. The models are either trained on CIFAR100 only (C), or continually fine-tuned on CIFAR100 after being trained on ImageNet100 (I2C).}
\label{exp: more I2C}
\end{table*}

\begin{table}[h]
\small
\setlength{\tabcolsep}{2.5pt}
\centering
\begin{tabular}{ccc}
\hline
Prompt & Method        &  I+C \\ \hline
$v+t$ &  S-liPrompts~\cite{wang2022s}                          &   39.6                          \\
\cline{1-1}
\arrayrulecolor{gray}\cline{1-1}\arrayrulecolor{black}
\multirow{4}{*}{$visual$} & \cellcolor{lgray}L2P-2~\cite{wang2022learning}                            & \cellcolor{lgray}41.6                          \\
& \cellcolor{lgray}DualPrompt-2~\cite{wang2022dualprompt}                       & \cellcolor{lgray}45.8                         \\
&  \cellcolor{lgray}CODA-P-2~\cite{smith2023coda}& \cellcolor{lgray}62.7   \\
& \cellcolor{lgray}HiDe-Prompt-2~\cite{wang2023hide}  & \cellcolor{lgray}75.3  \\
\cline{1-1}
\arrayrulecolor{gray}\cline{1-1}\arrayrulecolor{black}
\multirow{4}{*}{$text$}  &  \cellcolor{gray}CoOp-2~\cite{zhou2022learning} (1000)                         &  \cellcolor{gray}55.4                          \\
&  \cellcolor{gray}Continual-CLIP~\cite{thengane2022continualclip}                      &  \cellcolor{gray}65.9                          \\
&  \cellcolor{gray}Zero-Shot CLIP~\cite{radford2021learning}  &  \cellcolor{gray}68.7 \\
& \cellcolor{gray}AttriCLIP~\cite{wang2023attriclip}        &  \cellcolor{gray}\underline{78.3}                 \\
\midrule

$text^*$
& \textbf{ProTPS} (ours) & \textbf{85.6} \textcolor{darkgreen}{(+7.3)} \\

\bottomrule
\end{tabular}
\caption{Last accuracy of different methods on ImageNet100 + CIFAR100 (I+C) where each model is continually trained on ImageNet100 and CIFAR100 in sequence under CDC setting.}
\label{exp: more I+C}
\end{table}

APG~\cite{tang2023prompt} is also a visual-prompt-based method that demonstrates its effectiveness when training from scratch and has comparable performance to other existing visual-prompt-based methods when using the same pre-trained image encoder. Since APG~\cite{tang2023prompt} does not achieve the best performance among visual-prompt-based methods, we report its performance in this section with the same CLIP pre-trained ViT-L/14~\cite{dosovitskiy2020image} as the image encoder in Tab.~\ref{exp: SLCA & APG}. 

Besides, SLCA~\cite{zhang2023slca} is also built upon the pre-trained image encoder. But SLCA~\cite{zhang2023slca} is a non-prompt-based method that carefully finetunes the pre-trained image encoder with a smaller learning rate than in the classifier head. It models class-wise distributions in the form of a feature covariance matrix for every class and generates pseudo-training samples (features) of seen classes to align the classifier heads between consecutive tasks. Since SLCA~\cite{zhang2023slca} is not a prompt-based method, we report its performance in this section with the same CLIP pre-trained ViT-L/14~\cite{dosovitskiy2020image} as the image encoder in Tab.~\ref{exp: SLCA & APG}. $text^*$ denotes that the learning and selection of text prompts are instructed by vision prototypes in our ProTPS.

\subsection{More Results on Cross-Datasets Continual Learning}
 S-liPrompts~\cite{wang2022s} is a CLIP-based method, that uses both visual prompts and text prompts and derives the classifier via the text encoder. Yet, it is designed for domain incremental learning. By treating each dataset as a domain, we also test S-liPrompts~\cite{wang2022s} on the Cross-Datasets Continual Learning setting. The results are in Tab.~\ref{exp: more I2C} and Tab.~\ref{exp: more I+C}. $text^*$ denotes the learning and selection of text prompts are instructed by vision prototypes in our ProTPS. ``$v+t$'' denotes both visual and text prompts.

        

\subsection{SupCon Loss Details}
\label{sec: supcon}
In Tab.~\ref{exp: loss_cc}, we compare our proposed ``prompt-prototype contrastive loss'' and SupCon~\cite{khosla2020supervised} which is originally proposed to learn good encoders, i.e., representation learning. SupCon~\cite{khosla2020supervised} loss is defined within a batch of images as:
\begin{equation}
\mathcal{L}^{sup}_{out} = \sum_{i \in I} -\frac{1}{|P(i)|} \sum_{p \in P(i)} \log \frac{\exp (z_i \cdot z_p / \tau)}{\sum_{a \in A(i)} \exp (z_i \cdot z_a / \tau)},
\end{equation}
where $i$ is the image index in a batch, $P(i)$ is the set of indices of all positives of $ith$ image in this batch distinct from $i$. $A(i)$ is the set of indices of all images in this batch except $i$, $z_i$ represents the image feature of $ith$ image, ``$\cdot$'' denotes the inner product, and $\tau$ is a scalar temperature parameter.

\begin{table}
    \small
    \centering
    \begin{tabular}{cc}
        \toprule
        Loss function  & Last acc. \\
        \midrule
        $Loss_{pp}$ (ours) & \textbf{83.28} \\
        SupCon Loss~\cite{khosla2020supervised} & 82.57 \textcolor{lightred}{(-0.71)}\\
        \bottomrule
    \end{tabular}
    \captionof{table}{Different loss functions on CIFAR100~\cite{krizhevsky2009learning}.}
    \label{exp: loss_cc}
\end{table}

We adopt it with the following equation to replace our proposed ``prompt-prototype contrastive loss'' and report the performance in Tab.~\ref{exp: loss_cc}. Given a single image input of class $i$, there is only one positive pair between our vision prototypes and weight vectors of $classifier_2$ denoted as the yellow grid in the matrix shown in Fig.~\ref{fig:piepline}, thus the adopted SupCon loss is as follows:
\begin{equation}
\mathcal{L} = -\log \frac{\exp (I_i \cdot T_i / \tau)}{\sum_{\substack{j \neq i \text{ or } k \neq i}} \exp (I_j \cdot T_k / \tau)},
\end{equation}
where the numerator is the positive pair and the denominator includes all negative pairs between our vision prototypes and weight vectors of $classifier_2$ in our ProTPS.


\subsection{ProTPS's Upper Bound Accuracy}
We further test our ProTPS's upper bound accuracy by using all tasks' training data at the same time, the same as standard supervised training. The performance is reported in Tab.~\ref{table: ProTPS upper bound}. We can see that ProTPS can achieve better performance than Linear Probe CLIP which trains a linear classifier based on the CLIP image encoder.

\begin{table}[htb]
\small
    \setlength{\tabcolsep}{3pt}
    \centering
    \begin{tabular}{ccc}
        \toprule
        Method & CIFAR100 & ImageNet100 \\
        \hline
        Linear Probe CLIP & 86.69  & 95.52   \\
        \textbf{ProTPS} (ours) & \textbf{87.19} & \textbf{95.86} \\
        \bottomrule
    \end{tabular}
    \caption{Upper bound accuracy of `Linear Probe CLIP'~\cite{radford2021learning} and our ProTPS on CIFAR100~\cite{krizhevsky2009learning} and ImageNet100~\cite{deng2009imagenet}.}
    \label{table: ProTPS upper bound}
\end{table}

\subsection{Ablation on ProTPS's prototypes training strategy}
\begin{figure}[htbp]
    \centering
    \includegraphics[width=0.98\columnwidth]{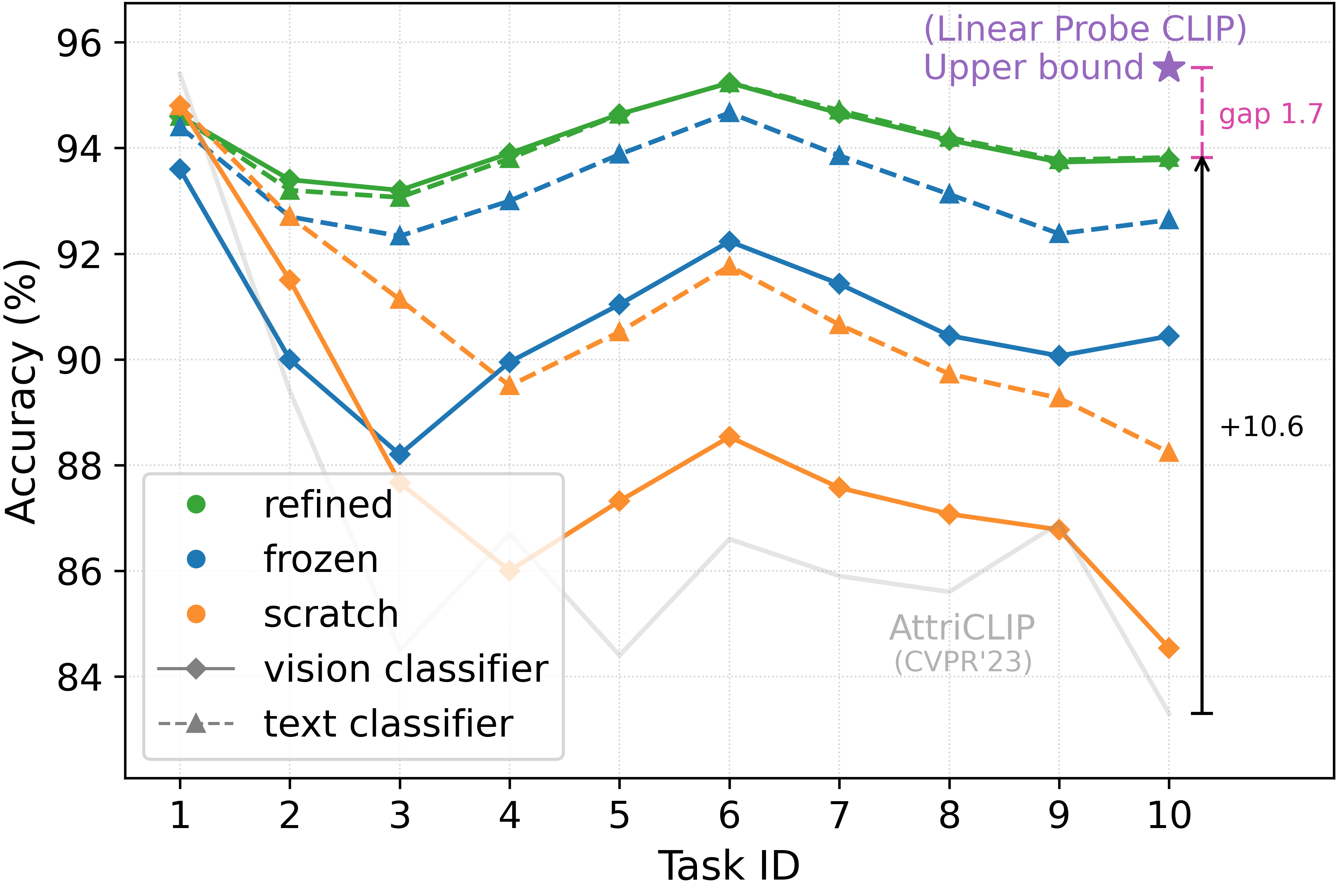}
    \caption{Ablation study of ProTPS's on ImageNet100~\cite{deng2009imagenet} under CI setting. The ``vision classifier'' and ``text classifier'' of ProTPS are well-aligned when using the proposed prototype learning strategy.}
    \label{fig:ablation_refine_imagenet100}
\end{figure}

The proposed prototype learning strategy is denoted as ``refined'' in Fig.~\ref{fig:ablation_refine_imagenet100}. We can see it is critical to help dual classifiers' alignment and achieves the best performance over training from ``scratch'' or  ``freezing'' them after the proposed initialization method.

\subsection{Pilot Study}
\label{sec: pilot}
\textbf{Inspiration.} Inspired by the effectiveness of using vision query for a different task~\cite{mei2024slvp}, we did the pilot study on CIFAR100 with ViT-L/14~\cite{dosovitskiy2020image} and ViT-B/16~\cite{dosovitskiy2020image} backbones in Tab.~\ref{full_pilot_study} for the class incremental learning task. Zero-shot CLIP's 18 prompt templates for CIFAR100 can be found in its official github. On both backbones, we consistently find that applying the basic nearest-prototype classifier to CLIP, denoted as Prototypes-CLIP, achieves slightly better performance than text prompt engineering and ensembling. This pilot study motivates us to use prototypes as guidance for our text prompt selection.

\begin{table*}[t]
    \centering
    \begin{tabular}{cccccccccccccc}
        \toprule
        Method & Classifier   & T-1 & T-2 & T-3 & T-4 & T-5 & T-6 & T-7 & T-8 & T-9 & T-10 \\
        \midrule
        Continual-CLIP~\cite{thengane2022continualclip} & text   & 94.7 & 88.5 & 76.8 & 73.6 & 72.4 & 72.8 & 69.6 & 68.5 & 68.1 & 66.7 \\
        Zero-Shot CLIP~\cite{radford2021learning} & text  & 95.6 & 89.2 & 77.7 & 74.6 & 73.9 & 74.1 & 70.6 & 70.1 & 69.6 & 68.3 \\
        Prototypes-CLIP & vision   & \textbf{95.6} & \textbf{90.7} & \textbf{81.5} & \textbf{77.1} & \textbf{75.9} & \textbf{75.7} & \textbf{72.8} & \textbf{72.3} & \textbf{71.4} & \textbf{69.9} \\
        \grayline
        Continual-CLIP~\cite{thengane2022continualclip} & text   & 95.3 & 92.3 & 82.1 & 79.2 & 78.5 & 77.9 & 75.7 & 75.2 & 74.8 & 73.4 \\
        Zero-Shot CLIP~\cite{radford2021learning} & text   & 95.7 & \textbf{92.9} & 85.2 & 82.9 & 82.3 & 82.1 & 79.6 & 79.6 & 79.2 & 78.3 \\
        Prototypes-CLIP & vision   & \textbf{97.6} & 92.8 & \textbf{88.1} & \textbf{85.3} & \textbf{84.3} & \textbf{84.3} & \textbf{81.9} & \textbf{81.9} & \textbf{81.2} & \textbf{80.0} \\
        \midrule
        Upper-bound (B/16) & -   & - & - & - & - & - & - & - & - & - & 82.5 \\
        Upper-bound (L/14) & -  & - & - & - & - & - & - & - & - & - & 86.7 \\
        \bottomrule
    \end{tabular}
    \caption{\textbf{Pilot Study.} Comparison of different CLIP-based methods in CI setting on CIFAR100~\cite{krizhevsky2009learning} with ViT-B/16~\cite{dosovitskiy2020image} (first 3 rows) or ViT-L/14~\cite{dosovitskiy2020image} (second 3 rows) backbone. After learning each task $T-i$, the accuracy on exposed classes from all the trained tasks (i.e., $T-1, 2, ..., t$) is reported. The upper-bound method is `Linear Probe CLIP'~\cite{radford2021learning} using all tasks' training data at the same time to optimize a linear classifier. The column `Classifier' denotes the encoder from which the classifier weights are derived.}
    \label{full_pilot_study}
\end{table*}

\section{More Text Prompts Visualizations}
\label{sec: CLIP templates}
\begin{figure*}[h]
    \centering
    \begin{subfigure}{0.15\linewidth} 
        \centering
        \includegraphics[width=0.5\linewidth]{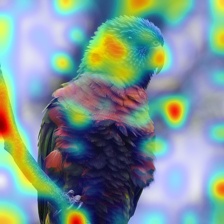}
    \end{subfigure}\hspace{-10mm}
    \begin{subfigure}{0.15\linewidth} 
        \centering
        \includegraphics[width=0.5\linewidth]{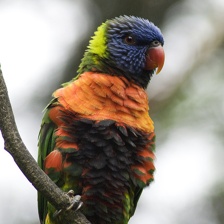}
    \end{subfigure}\hspace{-10mm}
    \begin{subfigure}{0.15\linewidth} 
        \centering
        \includegraphics[width=0.5\linewidth]{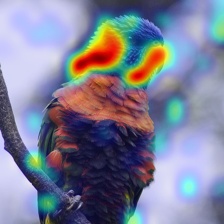}
    \end{subfigure}\hspace{-10mm}
    \hspace{10mm}
    \begin{subfigure}{0.15\linewidth} 
        \centering
        \includegraphics[width=0.5\linewidth]{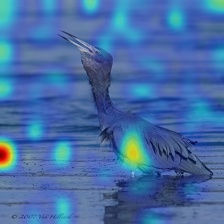}
    \end{subfigure}\hspace{-10mm}
    \begin{subfigure}{0.15\linewidth} 
        \centering
        \includegraphics[width=0.5\linewidth]{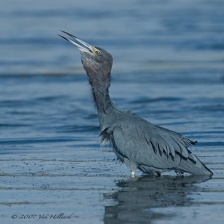}
    \end{subfigure}\hspace{-10mm}
    \begin{subfigure}{0.15\linewidth} 
        \centering
        \includegraphics[width=0.5\linewidth]{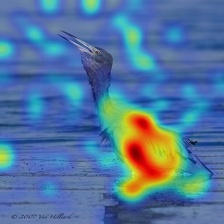}
    \end{subfigure}\hspace{-10mm}\\
    \vspace{0.5mm}
    \begin{subfigure}{0.15\linewidth} 
        \centering
        \includegraphics[width=0.5\linewidth]{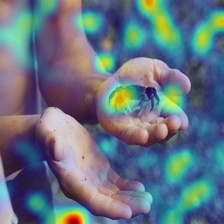}
    \end{subfigure}\hspace{-10mm}
    \begin{subfigure}{0.15\linewidth} 
        \centering
        \includegraphics[width=0.5\linewidth]{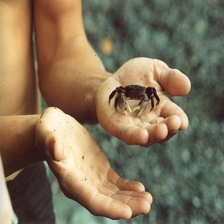}
    \end{subfigure}\hspace{-10mm}
    \begin{subfigure}{0.15\linewidth} 
        \centering
        \includegraphics[width=0.5\linewidth]{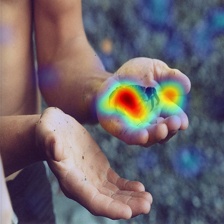}
    \end{subfigure}\hspace{-10mm}
    \hspace{10mm}
    \begin{subfigure}{0.15\linewidth} 
        \centering
        \includegraphics[width=0.5\linewidth]{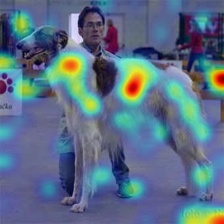}
    \end{subfigure}\hspace{-10mm}
    \begin{subfigure}{0.15\linewidth} 
        \centering
        \includegraphics[width=0.5\linewidth]{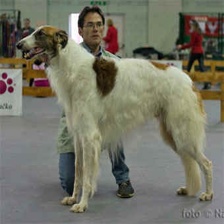}
    \end{subfigure}\hspace{-10mm}
    \begin{subfigure}{0.15\linewidth} 
        \centering
        \includegraphics[width=0.5\linewidth]{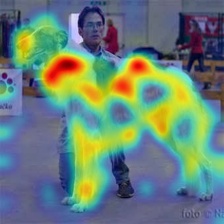}
    \end{subfigure}\hspace{-10mm}\\
    \vspace{0.5mm}
    \begin{subfigure}{0.15\linewidth} 
        \centering
        \includegraphics[width=0.5\linewidth]{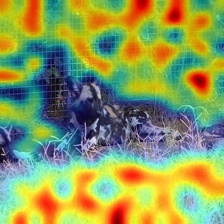}
    \end{subfigure}\hspace{-10mm}
    \begin{subfigure}{0.15\linewidth} 
        \centering
        \includegraphics[width=0.5\linewidth]{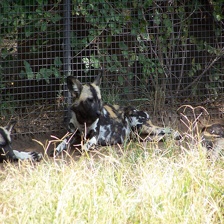}
    \end{subfigure}\hspace{-10mm}
    \begin{subfigure}{0.15\linewidth} 
        \centering
        \includegraphics[width=0.5\linewidth]{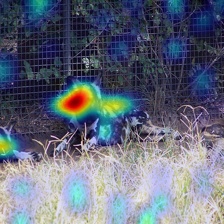}
    \end{subfigure}\hspace{-10mm}
    \hspace{10mm}
    \begin{subfigure}{0.15\linewidth} 
        \centering
        \includegraphics[width=0.5\linewidth]{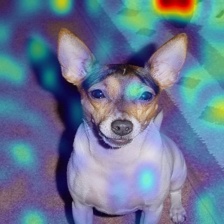}
    \end{subfigure}\hspace{-10mm}
    \begin{subfigure}{0.15\linewidth} 
        \centering
        \includegraphics[width=0.5\linewidth]{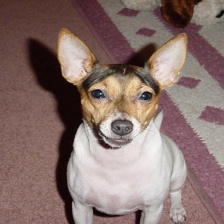}
    \end{subfigure}\hspace{-10mm}
    \begin{subfigure}{0.15\linewidth} 
        \centering
        \includegraphics[width=0.5\linewidth]{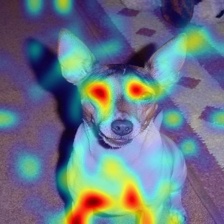}
    \end{subfigure}\hspace{-10mm}\\
    \vspace{0.5mm}
    \begin{subfigure}{0.15\linewidth} 
        \centering
        \includegraphics[width=0.5\linewidth]{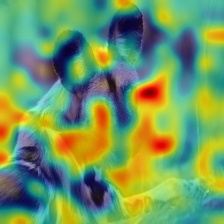}
    \end{subfigure}\hspace{-10mm}
    \begin{subfigure}{0.15\linewidth} 
        \centering
        \includegraphics[width=0.5\linewidth]{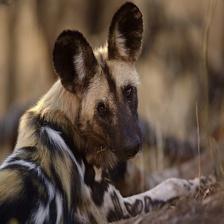}
    \end{subfigure}\hspace{-10mm}
    \begin{subfigure}{0.15\linewidth} 
        \centering
        \includegraphics[width=0.5\linewidth]{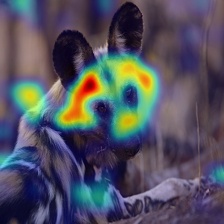}
    \end{subfigure}\hspace{-10mm}
    \hspace{10mm}
    \begin{subfigure}{0.15\linewidth} 
        \centering
        \includegraphics[width=0.5\linewidth]{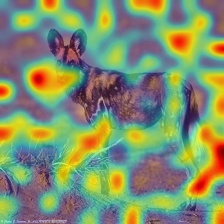}
    \end{subfigure}\hspace{-10mm}
    \begin{subfigure}{0.15\linewidth} 
        \centering
        \includegraphics[width=0.5\linewidth]{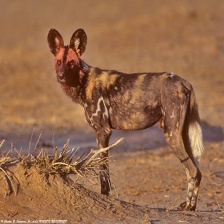}
    \end{subfigure}\hspace{-10mm}
    \begin{subfigure}{0.15\linewidth} 
        \centering
        \includegraphics[width=0.5\linewidth]{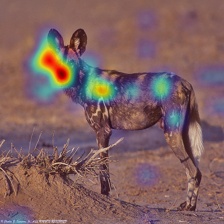}
    \end{subfigure}\hspace{-10mm}\\
    \vspace{0.5mm}
    \begin{subfigure}{0.15\linewidth} 
        \centering
        \includegraphics[width=0.5\linewidth]{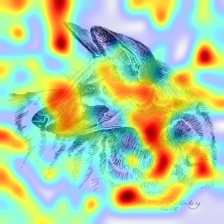}
    \end{subfigure}\hspace{-10mm}
    \begin{subfigure}{0.15\linewidth} 
        \centering
        \includegraphics[width=0.5\linewidth]{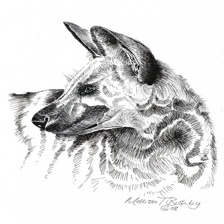}
    \end{subfigure}\hspace{-10mm}
    \begin{subfigure}{0.15\linewidth} 
        \centering
        \includegraphics[width=0.5\linewidth]{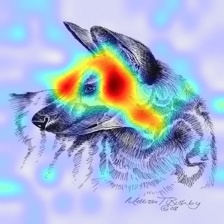}
    \end{subfigure}\hspace{-10mm}
    \hspace{10mm}
    \begin{subfigure}{0.15\linewidth} 
        \centering
        \includegraphics[width=0.5\linewidth]{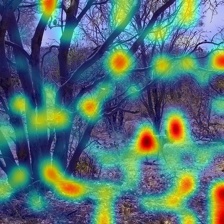}
    \end{subfigure}\hspace{-10mm}
    \begin{subfigure}{0.15\linewidth} 
        \centering
        \includegraphics[width=0.5\linewidth]{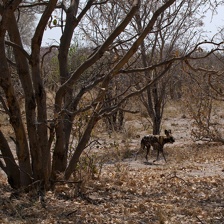}
    \end{subfigure}\hspace{-10mm}
    \begin{subfigure}{0.15\linewidth} 
        \centering
        \includegraphics[width=0.5\linewidth]{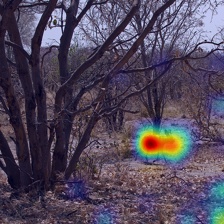}
    \end{subfigure}\hspace{-10mm}\\
    \vspace{0.5mm}
    \begin{subfigure}{0.15\linewidth} 
        \centering
        \includegraphics[width=0.5\linewidth]{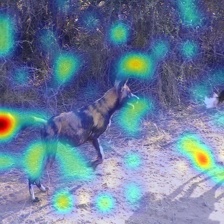}
    \end{subfigure}\hspace{-10mm}
    \begin{subfigure}{0.15\linewidth} 
        \centering
        \includegraphics[width=0.5\linewidth]{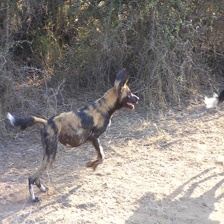}
    \end{subfigure}\hspace{-10mm}
    \begin{subfigure}{0.15\linewidth} 
        \centering
        \includegraphics[width=0.5\linewidth]{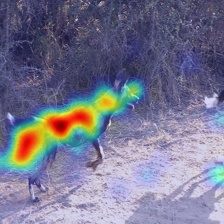}
    \end{subfigure}\hspace{-10mm}
    \hspace{10mm}
    \begin{subfigure}{0.15\linewidth} 
        \centering
        \includegraphics[width=0.5\linewidth]{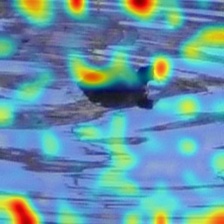}
    \end{subfigure}\hspace{-10mm}
    \begin{subfigure}{0.15\linewidth} 
        \centering
        \includegraphics[width=0.5\linewidth]{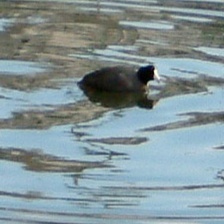}
    \end{subfigure}\hspace{-10mm}
    \begin{subfigure}{0.15\linewidth} 
        \centering
        \includegraphics[width=0.5\linewidth]{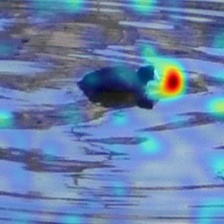}
    \end{subfigure}\hspace{-10mm}\\
    \vspace{0.5mm}
    \begin{subfigure}{0.15\linewidth} 
        \centering
        \includegraphics[width=0.5\linewidth]{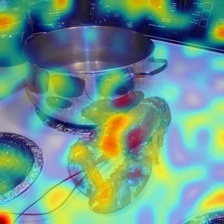}
    \end{subfigure}\hspace{-10mm}
    \begin{subfigure}{0.15\linewidth} 
        \centering
        \includegraphics[width=0.5\linewidth]{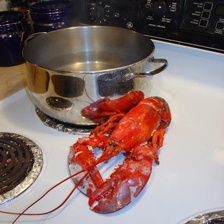}
    \end{subfigure}\hspace{-10mm}
    \begin{subfigure}{0.15\linewidth} 
        \centering
        \includegraphics[width=0.5\linewidth]{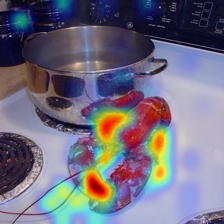}
    \end{subfigure}\hspace{-10mm}
    \hspace{10mm}
    \begin{subfigure}{0.15\linewidth} 
        \centering
        \includegraphics[width=0.5\linewidth]{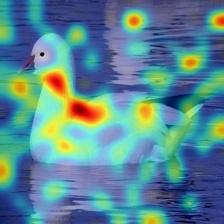}
    \end{subfigure}\hspace{-10mm}
    \begin{subfigure}{0.15\linewidth} 
        \centering
        \includegraphics[width=0.5\linewidth]{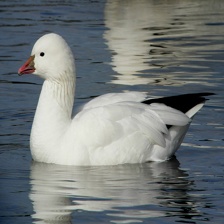}
    \end{subfigure}\hspace{-10mm}
    \begin{subfigure}{0.15\linewidth} 
        \centering
        \includegraphics[width=0.5\linewidth]{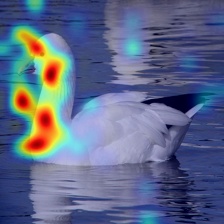}
    \end{subfigure}\hspace{-10mm}\\
    \vspace{0.5mm}
    \begin{subfigure}{0.15\linewidth} 
        \centering
        \includegraphics[width=0.5\linewidth]{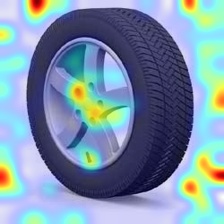}
    \end{subfigure}\hspace{-10mm}
    \begin{subfigure}{0.15\linewidth} 
        \centering
        \includegraphics[width=0.5\linewidth]{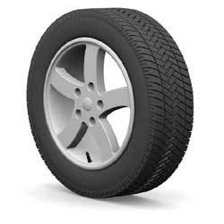}
    \end{subfigure}\hspace{-10mm}
    \begin{subfigure}{0.15\linewidth} 
        \centering
        \includegraphics[width=0.5\linewidth]{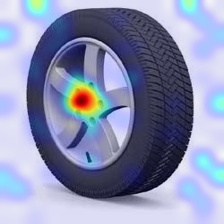}
    \end{subfigure}\hspace{-10mm}
    \hspace{10mm}
    \begin{subfigure}{0.15\linewidth} 
        \centering
        \includegraphics[width=0.5\linewidth]{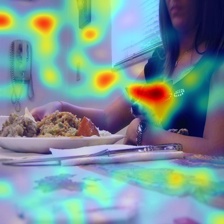}
    \end{subfigure}\hspace{-10mm}
    \begin{subfigure}{0.15\linewidth} 
        \centering
        \includegraphics[width=0.5\linewidth]{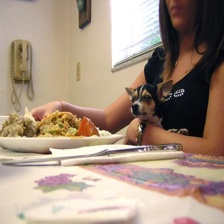}
    \end{subfigure}\hspace{-10mm}
    \begin{subfigure}{0.15\linewidth} 
        \centering
        \includegraphics[width=0.5\linewidth]{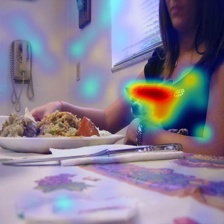}
    \end{subfigure}\hspace{-10mm}\\
    \vspace{0.5mm}
    \begin{subfigure}{0.15\linewidth} 
        \centering
        \includegraphics[width=0.5\linewidth]{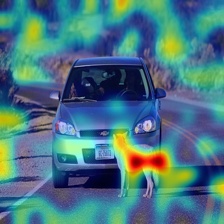}
    \end{subfigure}\hspace{-10mm}
    \begin{subfigure}{0.15\linewidth} 
        \centering
        \includegraphics[width=0.5\linewidth]{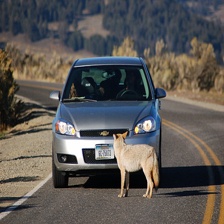}
    \end{subfigure}\hspace{-10mm}
    \begin{subfigure}{0.15\linewidth} 
        \centering
        \includegraphics[width=0.5\linewidth]{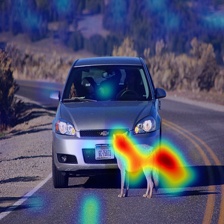}
    \end{subfigure}\hspace{-10mm}
    \hspace{10mm}
    \begin{subfigure}{0.15\linewidth}
        \centering
        \includegraphics[width=0.5\linewidth]{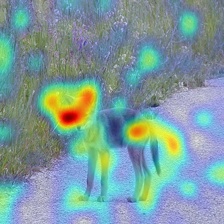}
    \end{subfigure}\hspace{-10mm}
    \begin{subfigure}{0.15\linewidth}
        \centering
        \includegraphics[width=0.5\linewidth]{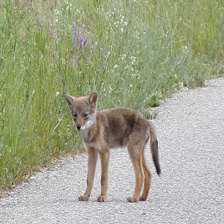}
    \end{subfigure}\hspace{-10mm}
    \begin{subfigure}{0.15\linewidth}
        \centering
        \includegraphics[width=0.5\linewidth]{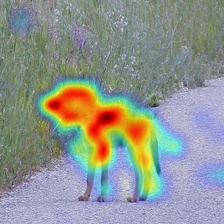}
    \end{subfigure}\hspace{-10mm}\\
    \vspace{0.5mm}
    \begin{subfigure}{0.15\linewidth}
        \centering
        \includegraphics[width=0.5\linewidth]{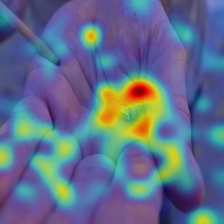}
    \end{subfigure}\hspace{-10mm}
    \begin{subfigure}{0.15\linewidth}
        \centering
        \includegraphics[width=0.5\linewidth]{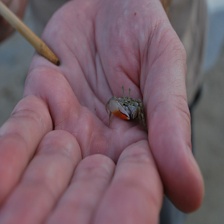}
    \end{subfigure}\hspace{-10mm}
    \begin{subfigure}{0.15\linewidth}
        \centering
        \includegraphics[width=0.5\linewidth]{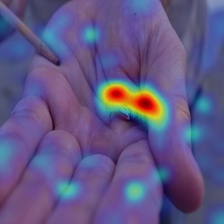}
    \end{subfigure}\hspace{-10mm}
    \hspace{10mm}
    \begin{subfigure}{0.15\linewidth}
        \centering
        \includegraphics[width=0.5\linewidth]{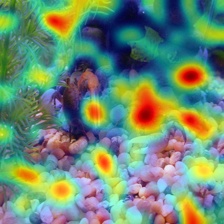}
    \end{subfigure}\hspace{-10mm}
    \begin{subfigure}{0.15\linewidth}
        \centering
        \includegraphics[width=0.5\linewidth]{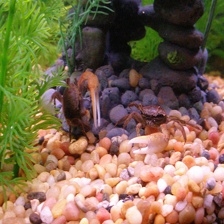}
    \end{subfigure}\hspace{-10mm}
    \begin{subfigure}{0.15\linewidth}
        \centering
        \includegraphics[width=0.5\linewidth]{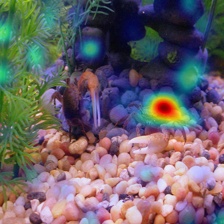}
    \end{subfigure}\hspace{-10mm}\\
    \vspace{0.5mm}
    \begin{subfigure}{0.15\linewidth}
        \centering
        \includegraphics[width=0.5\linewidth]{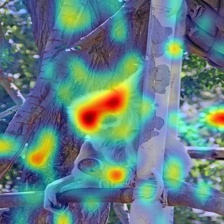}
    \end{subfigure}\hspace{-10mm}
    \begin{subfigure}{0.15\linewidth}
        \centering
        \includegraphics[width=0.5\linewidth]{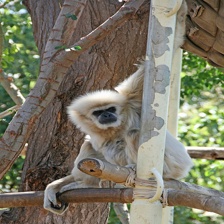}
    \end{subfigure}\hspace{-10mm}
    \begin{subfigure}{0.15\linewidth}
        \centering
        \includegraphics[width=0.5\linewidth]{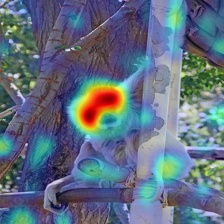}
    \end{subfigure}\hspace{-10mm}
    \hspace{10mm}
    \begin{subfigure}{0.15\linewidth}
        \centering
        \includegraphics[width=0.5\linewidth]{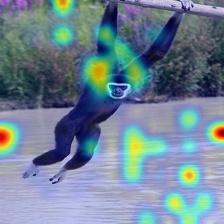}
    \end{subfigure}\hspace{-10mm}
    \begin{subfigure}{0.15\linewidth}
        \centering
        \includegraphics[width=0.5\linewidth]{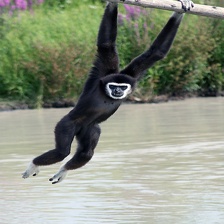}
    \end{subfigure}\hspace{-10mm}
    \begin{subfigure}{0.15\linewidth}
        \centering
        \includegraphics[width=0.5\linewidth]{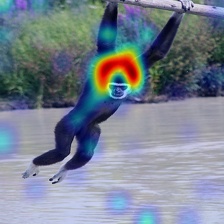}
    \end{subfigure}\hspace{-10mm}\\
    \vspace{0.5mm}
    \begin{subfigure}{0.15\linewidth}
        \centering
        \includegraphics[width=0.5\linewidth]{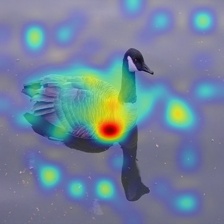}
    \end{subfigure}\hspace{-10mm}
    \begin{subfigure}{0.15\linewidth}
        \centering
        \includegraphics[width=0.5\linewidth]{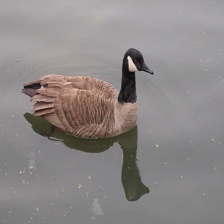}
    \end{subfigure}\hspace{-10mm}
    \begin{subfigure}{0.15\linewidth}
        \centering
        \includegraphics[width=0.5\linewidth]{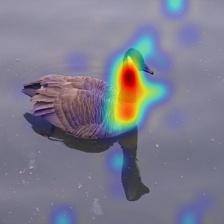}
    \end{subfigure}\hspace{-10mm}
    \hspace{10mm}
    \begin{subfigure}{0.15\linewidth}
        \centering
        \includegraphics[width=0.5\linewidth]{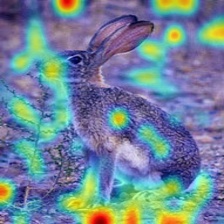}
    \end{subfigure}\hspace{-10mm}
    \begin{subfigure}{0.15\linewidth}
        \centering
        \includegraphics[width=0.5\linewidth]{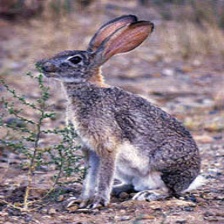}
    \end{subfigure}\hspace{-10mm}
    \begin{subfigure}{0.15\linewidth}
        \centering
        \includegraphics[width=0.5\linewidth]{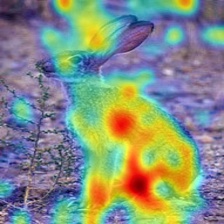}
    \end{subfigure}\hspace{-10mm}\\
    \vspace{0.5mm}
    \begin{subfigure}{0.15\linewidth}
        \centering
        \includegraphics[width=0.5\linewidth]{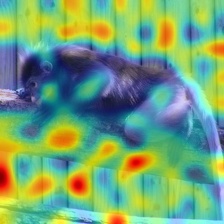}
    \end{subfigure}\hspace{-10mm}
    \begin{subfigure}{0.15\linewidth}
        \centering
        \includegraphics[width=0.5\linewidth]{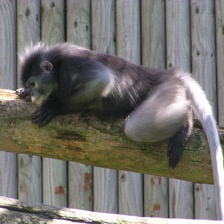}
    \end{subfigure}\hspace{-10mm}
    \begin{subfigure}{0.15\linewidth}
        \centering
        \includegraphics[width=0.5\linewidth]{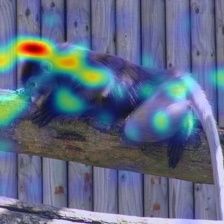}
    \end{subfigure}\hspace{-10mm}
    \hspace{10mm}
    \begin{subfigure}{0.15\linewidth}
        \centering
        \includegraphics[width=0.5\linewidth]{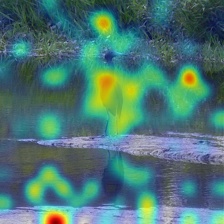}
    \end{subfigure}\hspace{-10mm}
    \begin{subfigure}{0.15\linewidth}
        \centering
        \includegraphics[width=0.5\linewidth]{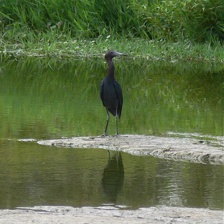}
    \end{subfigure}\hspace{-10mm}
    \begin{subfigure}{0.15\linewidth}
        \centering
        \includegraphics[width=0.5\linewidth]{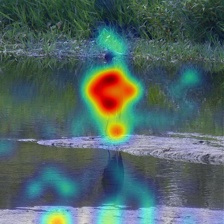}
    \end{subfigure}\hspace{-10mm}\\
    \vspace{0.5mm}
    \begin{subfigure}{0.15\linewidth}
        \centering
        \includegraphics[width=0.5\linewidth]{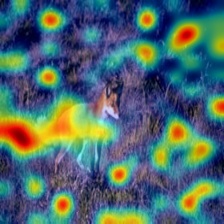}
    \end{subfigure}\hspace{-10mm}
    \begin{subfigure}{0.15\linewidth}
        \centering
        \includegraphics[width=0.5\linewidth]{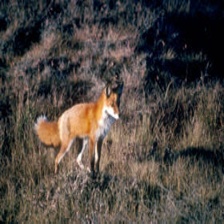}
    \end{subfigure}\hspace{-10mm}
    \begin{subfigure}{0.15\linewidth}
        \centering
        \includegraphics[width=0.5\linewidth]{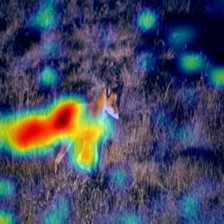}
    \end{subfigure}\hspace{-10mm}
    \hspace{10mm}
    \begin{subfigure}{0.15\linewidth}
        \centering
        \includegraphics[width=0.5\linewidth]{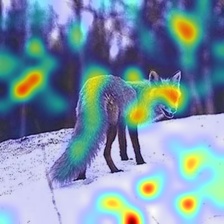}
    \end{subfigure}\hspace{-10mm}
    \begin{subfigure}{0.15\linewidth}
        \centering
        \includegraphics[width=0.5\linewidth]{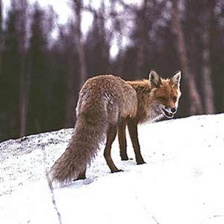}
    \end{subfigure}\hspace{-10mm}
    \begin{subfigure}{0.15\linewidth}
        \centering
        \includegraphics[width=0.5\linewidth]{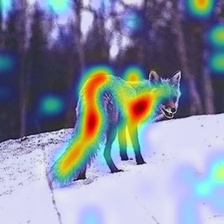}
    \end{subfigure}\hspace{-10mm}\\
    \vspace{0.5mm}
    \begin{subfigure}{0.15\linewidth}
        \centering
        \includegraphics[width=0.5\linewidth]{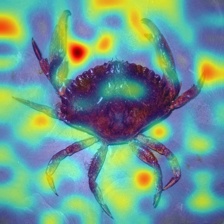}
    \end{subfigure}\hspace{-10mm}
    \begin{subfigure}{0.15\linewidth}
        \centering
        \includegraphics[width=0.5\linewidth]{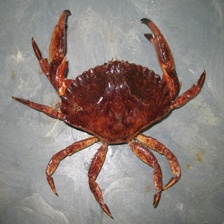}
    \end{subfigure}\hspace{-10mm}
    \begin{subfigure}{0.15\linewidth}
        \centering
        \includegraphics[width=0.5\linewidth]{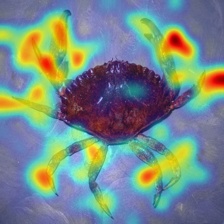}
    \end{subfigure}\hspace{-10mm}
    \hspace{10mm}
    \begin{subfigure}{0.15\linewidth}
        \centering
        \includegraphics[width=0.5\linewidth]{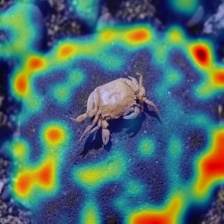}
    \end{subfigure}\hspace{-10mm}
    \begin{subfigure}{0.15\linewidth}
        \centering
        \includegraphics[width=0.5\linewidth]{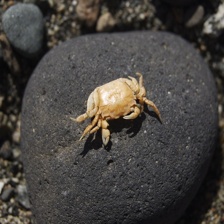}
    \end{subfigure}\hspace{-10mm}
    \begin{subfigure}{0.15\linewidth}
        \centering
        \includegraphics[width=0.5\linewidth]{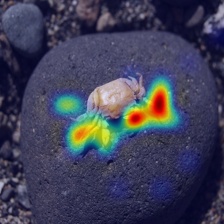}
    \end{subfigure}\hspace{-10mm}\\
    \vspace{0.5mm}
    \begin{subfigure}{0.15\linewidth}
        \centering
        \includegraphics[width=0.5\linewidth]{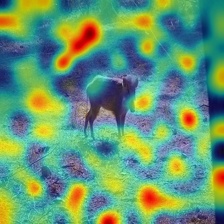}
    \end{subfigure}\hspace{-10mm}
    \begin{subfigure}{0.15\linewidth}
        \centering
        \includegraphics[width=0.5\linewidth]{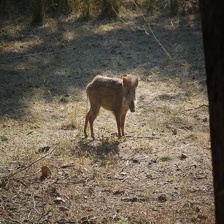}
    \end{subfigure}\hspace{-10mm}
    \begin{subfigure}{0.15\linewidth}
        \centering
        \includegraphics[width=0.5\linewidth]{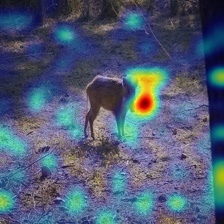}
    \end{subfigure}\hspace{-10mm}
    \hspace{10mm}
    \begin{subfigure}{0.15\linewidth}
        \centering
        \includegraphics[width=0.5\linewidth]{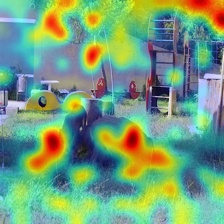}
    \end{subfigure}\hspace{-10mm}
    \begin{subfigure}{0.15\linewidth}
        \centering
        \includegraphics[width=0.5\linewidth]{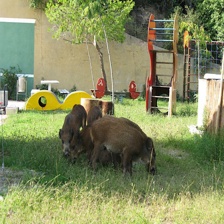}
    \end{subfigure}\hspace{-10mm}
    \begin{subfigure}{0.15\linewidth}
        \centering
        \includegraphics[width=0.5\linewidth]{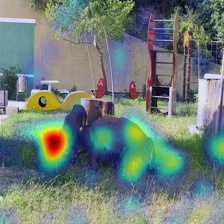}
    \end{subfigure}\hspace{-10mm}\\

    \caption{1st and 4th column: CLIP. 2nd and 5th column: Image. 3rd and 6th column: Ours. Text prompts visualized via Grad-CAM~\cite{selvaraju2017grad} on randomly selected classes of ImageNet100.}
    \label{fig: more vis}
\end{figure*}

The ``CLIP'' column in Fig.~\ref{fig: prompt vis} shows the attention of the original CLIP's prompts on paired images on ImageNet100. The original CLIP's 80 prompt templates for ImageNet can be found in CLIP's official github. The way to create ``zero-shot classifier weights'' for each class is also introduced in that link. For each class, we use the corresponding ``zero-shot classifier weights'' for visualization of the CLIP column in Fig.~\ref{fig: prompt vis}. More visualizations of text prompts from randomly selected classes are shown in Fig.~\ref{fig: more vis}. We can see that our learned unique text prompts can attend to region-level image details.



